\documentclass[runningheads]{llncs}

\usepackage{eccv}

\usepackage{eccvabbrv}

\usepackage{multirow}
\usepackage{graphicx}
\usepackage{rotating}
\usepackage[export]{adjustbox}
\usepackage{booktabs}
\usepackage[inline]{enumitem}
\usepackage{stackengine}
\usepackage{xcolor}

\usepackage[accsupp]{axessibility}  %

\usepackage[pagebackref,breaklinks,colorlinks,citecolor=eccvblue]{hyperref}

\usepackage{orcidlink}

\begin{document}

\newcommand{\OURS}{Lookalike3D}

\newcommand{\id}{\mathsf{Id}}
\newcommand{\simi}{\mathsf{Sim}}
\newcommand{\diff}{\mathsf{Diff}}

\newcommand{\DatasetName}{3DTwins}
\newcommand{\DatasetNumScenes}{906}
\newcommand{\DatasetNumTotalPairs}{76k}
\newcommand{\DatasetNumIdenticalPairs}{34.5k}
\newcommand{\DatasetNumSimilarPairs}{6k}
\newcommand{\DatasetNumDifferentPairs}{35.5}

\newcommand{\ImprovementGT}{104}

\newcommand{\ImprovementGTAbs}{0.25}
\newcommand{\ImprovementPredAbs}{0.13}

\title{\OURS{}: Seeing Double in 3D}

\author{Chandan Yeshwanth \quad Angela Dai}

\authorrunning{C. Yeshwanth et al.}

\institute{Technical University of Munich}

\makeatletter
\let\@oldmaketitle\@maketitle
\renewcommand{\@maketitle}{
  \@oldmaketitle
  \centering
  \includegraphics[width=\textwidth]{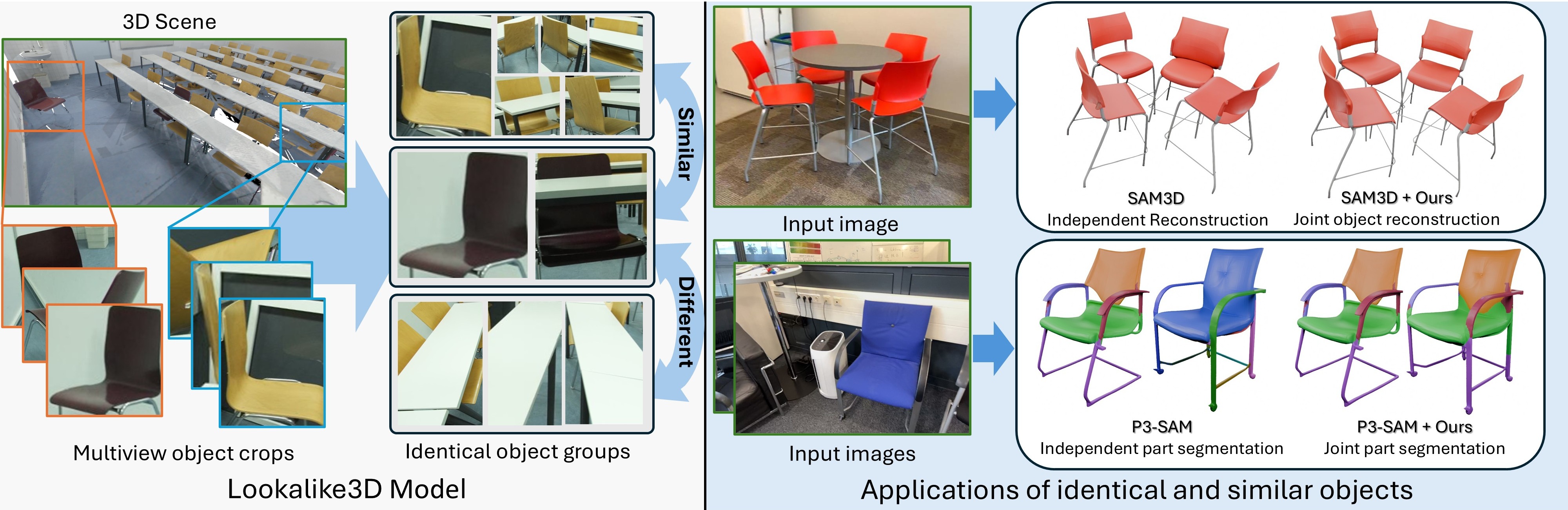}
  \captionof{figure}{Many real-world scenes naturally contain repeated objects.
  \OURS{} proposes to identify such groups of identical or near-identical (similar) objects across multiview images of an indoor scene. This enables holistic reasoning to be exploited in object-focused 3D reconstruction and perception methods. In particular, we demonstrate joint optimization under these global scene constraints using state-of-the-art, pre-trained 3D object reconstruction and part segmentation models. 
  }
  \label{fig:teaser}
  \bigskip
}
\makeatother

\maketitle

\begin{abstract}
  3D object understanding and generation methods produce impressive results, yet they often overlook a pervasive source of information in real-world scenes: repeated objects.
  We introduce the task of lookalike object detection in indoor scenes, which leverages repeated and complementary cues from identical and near-identical object pairs. Given an input scene, the task is to classify pairs of objects as identical, similar or different using multiview images as input. To address this, we present \OURS{}, a multiview image transformer that effectively distinguishes such object pairs by harnessing strong semantic priors from large image foundation models.
  To support this task, we collected the \DatasetName{} dataset, containing \DatasetNumTotalPairs{} manually annotated identical, similar and different pairs of objects based on ScanNet++, and show an improvement of \ImprovementGT{}\% IoU over baselines. 
  We demonstrate how our method improves downstream tasks such as enabling joint 3D object reconstruction and part co-segmentation, turning repeated and lookalike objects into a powerful cue for consistent, high-quality 3D perception.
  Our code, dataset and  models will be made publicly available.
\end{abstract}

\section{Introduction}
\label{sec:intro}
3D scene understanding and generation are important building blocks of content creation, design, and virtual reality (VR) applications, all of which rely on clean and consistent 3D assets for fidelity. Recent work on object-level 3D generation and perception have made impressive strides; 
 for instance, 3D object generative methods can produce high-fidelity object-level assets from image inputs \cite{chen2025sam, xiang2025structured, xiang2025trellis2}. Other methods show excellent results in making 3D assets decomposable or interactable by inferring object part decomposition\cite{liu2025partfield,ma2025p3} and articulation \cite{chen2025freeart3d, li2025particulate,gao2025meshart}. 
However, these methods largely ignore a pervasive cue in real-world scenes: repeated objects.
Recognizing which objects are identical or near-identical provides rich, complementary information that can significantly improve reconstruction, part understanding, and semantic reasoning.

Current image-to-3D methods are most reliable when objects are fully visible. When objects are partially occluded, these models must hallucinate missing geometry, since they operate primarily on single objects with limited context. However, real indoor scenes often contain multiple instances of the same object, offering complementary visual perspectives.
For example, one chair may be seen from the front while another is seen from the side. Despite their ubiquity, repeated objects are rarely leveraged in 3D reconstruction pipelines.

We observe that object pairs in indoor scenes naturally fall into three categories:
\begin{enumerate*}[label=\arabic*)]
\item \textit{identical} pairs that have exactly identical shape and texture
\item \textit{similar} pairs: distinct objects that differ slightly in shape or texture
\item \textit{different} pairs that differ significantly either in shape or texture
\end{enumerate*}. \textit{Identical} and \textit{similar} pairs contain highly redundant information that can be exploited for joint reconstruction and semantic understanding.  Motivated by this, we introduce the task of \textit{lookalike object detection}: given input multiview images of objects in an indoor scene, the task is to classify each pair of objects within the same semantic class as identical, similar, or different.

To solve this task, we formulate lookalike detection as a similarity learning problem. Our method, \OURS{}, takes multiview images of object pairs and embeds them into a feature space where proximity reflects similarity. We leverage the strong semantic prior of image foundation models such as DINOv2 \cite{oquab2023dinov2} and propose an alternating attention \cite{wang2025vggt} model paired with a multi-class triplet loss and score alignment loss for similarity learning. Our model can effectively capture fine-grained appearance and geometry cues in multiview images which are not available in coarse 3D point cloud inputs.

We demonstrate the utility of \OURS{} on two downstream tasks: single-view joint 3D object reconstruction of identical objects from casual camera captures, and part co-segmentation of similar objects. In both cases, we show that jointly optimizing identical or similar pairs of objects produces more consistent and detailed reconstructions or segmentations compared to treating objects independently.

To enable this task, we collected \DatasetName{}, a large-scale dataset of identical and near-identical objects based on the ScanNet++ dataset~\cite{yeshwanth2023scannet++} of 3D indoor scans containing \DatasetNumTotalPairs{} object pairs spanning identical, similar, and different categories. This dataset enables training, benchmarking, and evaluation of lookalike object detection methods and provides a standard for object matching in 3D indoor scenes.

Our contributions are:
\begin{itemize}
    \item we introduce the task of lookalike object detection, a task to identify identical and similar object pairs in 3D indoor scenes.
    \item we present a multiview image transformer-based similarity learning model with alternating attention layers to solve this task.
    \item we provide a large scale dataset of identical, similar and different object pairs annotated in real world 3D scans to enable training and benchmarking.
    \item we demonstrate applications of our method on two downstream tasks: consistent 3D reconstruction of identical objects, and part co-segmentation of similar objects.
\end{itemize}

\section{Related Work}
\label{sec:related_work}

\subsection{2D Object Re-identification}
Re-identification is a widely explored computer vision task operating on RGB images.  Given a query image of an object, the task is to retrieve the closest images from a database which contain the same object, under varying viewpoint, lighting conditions and occlusions. Most re-identification research has focused on two categories: humans and vehicles. Commonly used benchmarks for vehicle re-identification are VeRi-776 \cite{liu2016deep} and PKU VehicleID \cite{liu2016deepvehicleid}, and methods have used self-supervised learning \cite{li2021self}, classification and triplet losses on multiview images \cite{meng2020parsing} or contrastive loss \cite{lu2025clip} among others. Human re-identification methods \cite{hermans2017defense, wojke2018deep} use benchmarks such as Market-1501 \cite{zheng2015scalable} and solve the task similarly. These methods are restricted to a single object class and hence do not generalize to diverse and long-tail objects in 3D scenes. 
Recent work such as VICP \cite{huang2025generalizable} has attempted to solve the problem of general object re-identification on a larger set of classes, by similarity learning and in-context prompting. 
However, these methods are limited to a fixed set of classes and do not utilize multi-view images. Unlike re-identification approaches that output ranked lists, we leverage multi-view data to classify object pairs directly as identical, similar, or different.

\subsection{3D Object Re-identification}
Analogous to the 2D setting, 3D re-identification methods aim to associate previously seen 3D instances with new ones, taking a point cloud or other 3D representation as input. 3D Semantic MapNet \cite{cartillier20243d} uses coarse object-level features to match objects in a one-to-one manner. Living Scenes \cite{zhu2024living} associates objects over time, while Rescene4D \cite{steiner2026rescene4d} segments instances in a consistent manner over time. RIO \cite{wald2019rio} addresses the related task of object re-localization in 3D scenes. These methods jointly reconstruct single object representations based on coarse semantics rather than fine-grained details. Furthermore, they produce only one-to-one mappings, preventing them from grouping identical or similar objects. In contrast, we use RGB images to group objects via fine-grained appearance and shape cues, leveraging 3D foundation models for joint reconstruction.

\subsection{2D Image Matching}
Another task related to ours is RGB image matching, which is more general than re-identification. Image matching methods such as SuperGlue\cite{sarlin2020superglue} and ASpanformer \cite{chen2022aspanformer} can compare objects in images by predicting sparse keypoints and then matching them, and methods such as Mast3r \cite{leroy2024grounding} and TLFR \cite{zhang2024telling} predict dense pixel matches between images. CroCo v2 \cite{weinzaepfel2023croco} trains a masked view completion model and uses it for stereo matching, and ZeroCo \cite{an2025cross} extends it to dense matching. These methods are trained to match multiple views of the same object and depend heavily on surrounding context. Furthermore, they often lack multi-view support and capture limited object-level detail. In contrast, \OURS{} matches different instances of identical and similar objects, is robust to background changes, and effectively utilizes multi-view inputs.
    
\subsection{3D Point Cloud Registration}
The 3D equivalent of 2D matching is point cloud registration, which aims to predict point-level matches between two 3D objects or scenes.
Methods such as Lepard \cite{li2022lepard}, Predator \cite{huang2021predator}, RTMM \cite{therien2024object} and REGTR \cite{yew2022regtr} are designed to register point clouds of the same object or a complete scene with partial overlap. These methods tend to match based on coarse semantics (chair arm, leg, back) and not on fine-grained appearance details and hence predict confident matches on non-identical objects (e.g., chair leg matching with another different type of chair's leg).

\section{Method}
\label{sec:method}

\begin{figure*}[t]
  \centering
  \includegraphics[width=\textwidth]{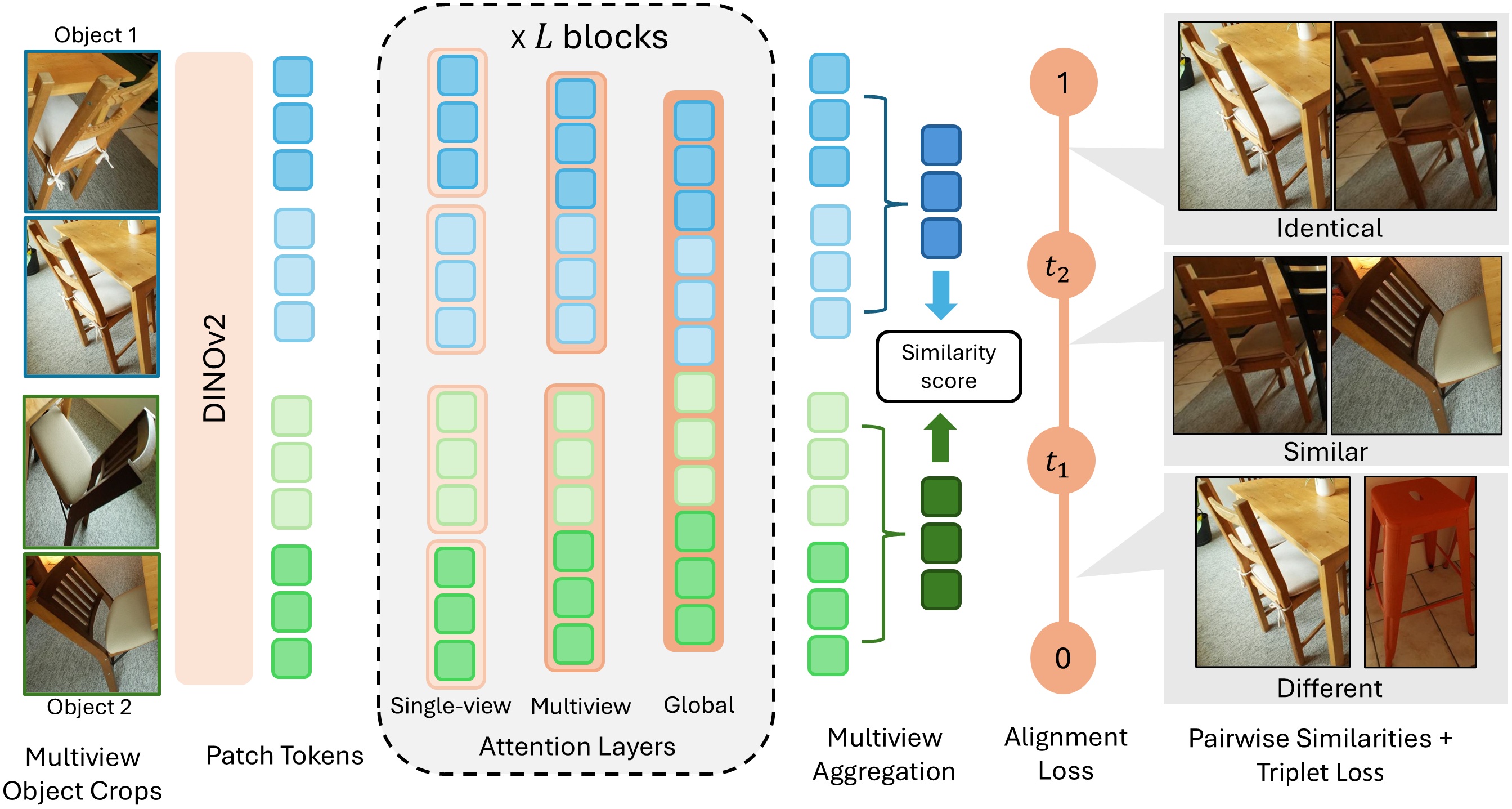}
  \caption{Overview of the \OURS{} model. \OURS{} takes as input a single image or multiple posed images each
  of a pair of objects, passes these through a shared frozen DINOv2 backbone to extract image patch features, and then encodes these patch features with three kinds of attention layers: single-view, multiview and global. The resulting patch features are aggregated and compared to obtain a single similarity measure of the two objects. The model is trained with a triplet loss to separate the identical, similar and different pairs, and an alignment loss to align the similarities to the classification thresholds.}
  \label{fig:architecture}
\end{figure*}

\paragraph{Problem Formulation.}
Indoor scenes often contain valuable information in the form of repeated object instances.
The task of lookalike object detection is to identify such groups of identical objects and pairs of similar objects in the scene. We take as input a sequence of RGB-D images containing a set of object instances $o_i = (m_i, \{I_{i,j}\})$ where $m_i$ is the binary mask of the object on $\mathcal{V}$, and $I_{i,j}$ are the multiview crops of the object $o_i$. The object instances can be obtained from ground truth data or off-the-shelf models that predict object masks on RGB images \cite{ravi2024sam} and aggregate them \cite{yan2024maskclustering}. RGB-D sequences are required only during training to associate the multiview images together; inference can be performed on a single RGB image.

The task is to predict groups of identical objects $\id_i = (o_{\id_{i,1}}, o_{\id_{i,2}}, \ldots)$ that have identical shape and appearance, and pairs of similar objects $\simi=((o_{\simi_{1,1}}, o_{\simi_{1,2}}),\ldots)$ that have near-identical shape or appearance. The remaining pairs of objects are classified into $\diff$ (different).
The RGB-D setting with a dense trajectory of images is particularly useful for this task as it allows for capturing both redundant and complementary information about objects, by observing them from different sides and fusing the results.

\paragraph{Overview of \OURS{}} First, \OURS{} creates pairs of objects $(o_1, o_2)$ in the scene and takes their image crops $I_{1,j}$ and $I_{2,j}$ as input, $j=1\ldots k$ and  $k$ is the maximum number of inputs views for an object. These images are embedded into patch features using a shared feature extractor \cite{oquab2023dinov2} to obtain patch features for each image. These patch features are then refined using a transformer encoder with self attention layers. The refined patch features are aggregated over all frames within an object, and the resulting features are compared across objects to compute similarity scores. These scores are used to classify the object pair into one of three categories: $\id{}$, $\simi{}$ and $\diff{}$. Object pairs in $\id{}$ share exact same shape and appearance, while object pairs in $\simi{}$ are near-identical. Detecting these two categories enables us to make use of rich information from repeated instances for various downstream applications.

\paragraph{Feature Extractor}
We use a shared DINOv2 backbone to extract patch features from the input image crops independently, resulting in $N_p$ patch features $p_{i, j}$ for image $I_{i, j}$. Along with the final layer features which encode high-level object and part semantics, we also extract intermediate features from layers $(1,3,5,8)$ to encode structural information of the objects \cite{amir2021deep}.

\vspace{-0.25cm}

\paragraph{Alternating Attention}
We utilize the transformer self-attention mechanism to process multiview images \cite{lin2025depth, wang2022matchformer, wang2025vggt}, and propose passing the  DINOv2 patch features $p_{i, j}^0$ through transformer encoder layers at three hierarchical levels: single-view, multiview (object) and global. Single-view layers $\Phi_{\mathrm{frame},l}$ apply self-attention to all tokens within a single object crop. Multiview (object-level) layers  $\Phi_{\mathrm{object},l}$ operate on all image tokens from the same object, and global layers $\Phi_{\mathrm{global},l}$ operate on all tokens across both objects. At layer $l$:
\vspace{-0.25cm}

\begin{align*}
    \text{Single-view:} \quad & p^{l}_{i,j} = \Phi_{\mathrm{frame},l}(p^{l-1}_{i,j}) \\
    \text{Multiview:} \quad & p^{l}_{i,j} = \Phi_{\mathrm{object},l}(\{p^{l}_{i,j}\}), \quad \forall j \in \{1, \dots, k\} \\
    \text{Global:} \quad & p^{l}_{i,j} = \Phi_{\mathrm{global},l}(\{p^{l}_{i,j}\}), \quad \forall i \in \{1, 2\}, \, \forall j \in \{1, \dots, k\}
\end{align*}

We add learned frame and object embeddings to the patch tokens in each layer to distinguish between tokens from different frames or different objects. We aggregate the patch features from the final layer across the input views for each object by averaging them to produce a single sequence of patch features for each object $f_{i}$. We then flatten the patch features, normalize them and compute the dot product of features of the two objects to obtain a single similarity score $s_{i,j} \in [0, 1]$ for objects $i$ and $j$. We observe from the global attention scores that the foreground region mask is learned accurately from the data; therefore, flattening the patch features does not negatively affect performance. The average operation allows \OURS{} to perform inference on a single RGB image or fewer views than used for training.

\paragraph{Loss functions} We use a combination of triplet and score alignment losses for similarity learning. The triplet loss optimizes the the relative similarity between identical, similar and different pairs, while the alignment loss ensures that similarity scores remain within defined thresholds. This prevents the need to tune thresholds for diverse and long-tail indoor scenes.

\paragraph{Triplet Loss} We use a triplet loss with margins \cite{schroff2015facenet} to disentangle the three categories of object pairs in the feature space: $\id{}$, $\simi{}$ and $\diff{}$. We achieve this by considering the different types of pairs in a pairwise manner. For a given anchor pair of ground truth (GT) identical objects, we encourage its similarity to be higher than that of corresponding similar or different object pairs sharing a common object with the identical pair, which results in two loss terms. The third loss term encourages similar pairs to be closer together in the feature space than different pairs. Since the triplet loss uses feature distances instead of similarities, we use the feature distance $d_{i,j} = 1 - s_{i,j}$. The combined loss is: 

\begin{align*}
    \mathcal{L_{\mathrm{triplet}}} = \sum_{a \in \mathcal{A}} & \max(0, d_{a, o_{\id}} - d_{a, o_{\simi}} + \alpha_1) \\
    + & \max(0, d_{a, o_{\simi}} - d_{a, o_{\diff}} + \alpha_2) \\
    + & \max(0, d_{a, o_{\id}} - d_{a, o_{\diff}} + \alpha_3)
\end{align*}

where  $\mathcal{A}$ is the set of anchor objects and $o_{\id}$, $o_{\simi}$ and $o_{\diff}$ represent ground truth identical, similar and different objects corresponding to the anchor object $a$, and $\alpha_i$ denote the margins between the respective classes. We select hard negatives for the triplet loss to efficiently distinguish the classes; for example, the first loss term compares the maximum $\id$ distance with the minimum $\simi$ distance.

\paragraph{Classification and Score Alignment Loss} While the triplet loss $\mathcal{L}$ separates $\id$, $\simi$ and $\diff$ in the feature space, we need to set a hard threshold to classify pairs into one of these three types. Hence, we use two thresholds $t_1$ and $t_2$ in the interval $[0, 1]$ such that 

\begin{align*}
\text{Label}(o_i, o_j) = 
\begin{cases} 
    \mathsf{Id}  & \text{if } s_{i,j} \in [t_2, 1] \\
    \mathsf{Sim} & \text{if } s_{i,j} \in [t_1, t_2) \\
    \mathsf{Diff} & \text{if } s_{i,j} \in [0, t_1)
\end{cases}
\end{align*}

To encourage the similarity scores to lie within these regions, we add a loss on the signed distance between the predicted similarity and the two ground truth thresholds for the identical and different bins:

\begin{align*}
\mathcal{L}_{\mathrm{align}} = \sum_{(o_i, o_j)} 
\begin{cases} 
    (t_2 - s_{i,j}) + (1 - s_{i,j}) & \text{if } (o_i, o_j) \in \mathsf{Id} \\
    (s_{i,j} - t_1) +  s_{i,j} & \text{if } (o_i, o_j) \in \mathsf{Diff}
\end{cases}
\end{align*}

Scores for similar pairs are aligned to the $\simi$ bin by the combination of  $\mathcal{L_{\mathrm{triplet}}}$ and $\mathcal{L}_{\mathrm{align}}$.
The overall loss is the sum of triplet and alignment losses $\mathcal{L} = \mathcal{L_{\mathrm{triplet}}} + \mathcal{L}_{\mathrm{align}}$. 
Finally, identical object pairs are transitively grouped together, while similar object pairs are kept as-is.

\paragraph{Implementation Details}
We set the classification thresholds $(t_1, t_2)=(0.33. 0.66)$ so that they are equally spaced within the similarity range $[0, 1]$. The top 5 views covering each object are selected. We use the Adam optimizer with a learning rate of $0.0001$. Training takes about 4 hours on an Nvidia A6000 GPU, and inference takes an average of $34$ms/sample with a batch size of 256. We use a DINOv2-small model which remains frozen throughout training. Our model consists of one alternating attention block, which contains 1 layer each of single-view, multiview and global attention. We set the margins $\alpha_i$ to $(0.4, 0.4, 0.8)$ for $\id$-$\simi$, $\simi$-$\diff$ and $\id$-$\diff$ respectively. 

\subsection{Applications of \OURS{}}
Our method produces groups of identical objects and pairs of similar objects in indoor scenes. We demonstrate the applications of these results on two downstream tasks: 3D object reconstruction and part co-segmentation. 

\vspace{-0.25cm}

\subsubsection{Joint 3D Object Reconstruction}
We use the identical objects detected by \OURS{} for the task of joint 3D object reconstruction. SAM 3D Objects \cite{chen2025sam} takes an input image and the corresponding object mask as input, and outputs a high-fidelity 3D textured mesh of the object. As it infers each object independently, it reconstructs slightly different versions of identical objects. We leverage the redundant information from repeated identical instances of an object to obtain a single coherent reconstruction.

We adapt SAM 3D to jointly reconstruct a single object from a group of identical instances. We provide a single image containing all instances along with multiple object masks as input. These masks are injected as conditions into every layer of the geometry and texture/refinement models.

We input a single image and $k$ masks of identical objects as conditions into the Mixture-of-Transformers (MoT) geometry model. To predict the $16^3$ occupancy features, we average tokens across the $k$ outputs after each layer and pass the result to the next layer. The model finally upsamples to $64^3$ occupancy features of the jointly reconstructed object, and predicts  a single rotation, scale and translation.

In the texture and refinement stage, we initialize sparse locations with the previously predicted joint  occupancy grid, and similarly utilize the object masks of $k$ identical objects as conditions for every layer of the flow transformer which is used to sample SLAT \cite{xiang2025structured} features. We then average over the intermediate outputs and use these as input to the next layer. The output of this model is a single set of SLAT features at sparse 3D locations corresponding to the joint model. These are passed through the pretrained VAE decoder to obtain the jointly reconstructed mesh of the $k$ input objects in canonical space. We transform this joint model to each of $k$ locations using the rotation and translation obtained through individual reconstructions. We use the median of the individual scales as the scale of the jointly reconstruction object.

\vspace{-0.25cm}

\subsubsection{Part Co-segmentation}
We demonstrate an application of predicted similar object pairs from \OURS{}. Because similar objects differ only slightly in shape or appearance, they share largely the same semantics. For example, similar chairs maintain comparable part structures, such as legs, armrests, seats, and backs.

P3-SAM \cite{ma2025p3} provides high-quality 3D part decomposition but may under-segment objects with ambiguous geometry, such as fusing a chair’s back and seat. We address this by transferring segments from a correctly segmented source object to an under-segmented target. We first generate the 3D objects corresponding to similar objects using SAM 3D independently on each object, and then align the 3D objects using point-to-point ICP; since SAM 3D produces models in canonical coordinates, alignment is typically stable. Potential ICP failures can be mitigated by running binned rotations about the vertical axis. Finally, we assign part IDs to the target object using a majority vote (mode) from the $k$-nearest neighbor points of the source object.

\section{Dataset}
\label{sec:dataset}

\begin{table}[ht]
\centering

\begin{tabular}{@{}lrrr@{}}
\toprule
\textbf{Category} & \textbf{Train} & \textbf{Val} & \textbf{Overall} \\ \midrule
\quad Identical Pairs & 32,270 & 2,496 & 34,766 \\
\quad Similar Pairs   & 5,468 & 735 & 6,203 \\
\quad Different Pairs & 33,320 & 2,450 & 35,770 \\
\midrule
\quad \textbf{Total Pairs} & \textbf{71,058} & \textbf{5,681} & \textbf{76,739} \\ \midrule
\quad Unique Object Instances & 22,285 & 1,563 & 23,848 \\
\quad Object Classes          & 867 & 148 & 1015 \\ \bottomrule
\end{tabular}
\vspace{0.2cm}
\caption{Dataset statistics for identical, similar, and different object pairs, and number of unique object instances and classes in the \DatasetName{} dataset.}
\label{tab:dataset_stats}
\end{table}

\begin{figure*}[t]
    \centering

    \includegraphics[width=\textwidth]{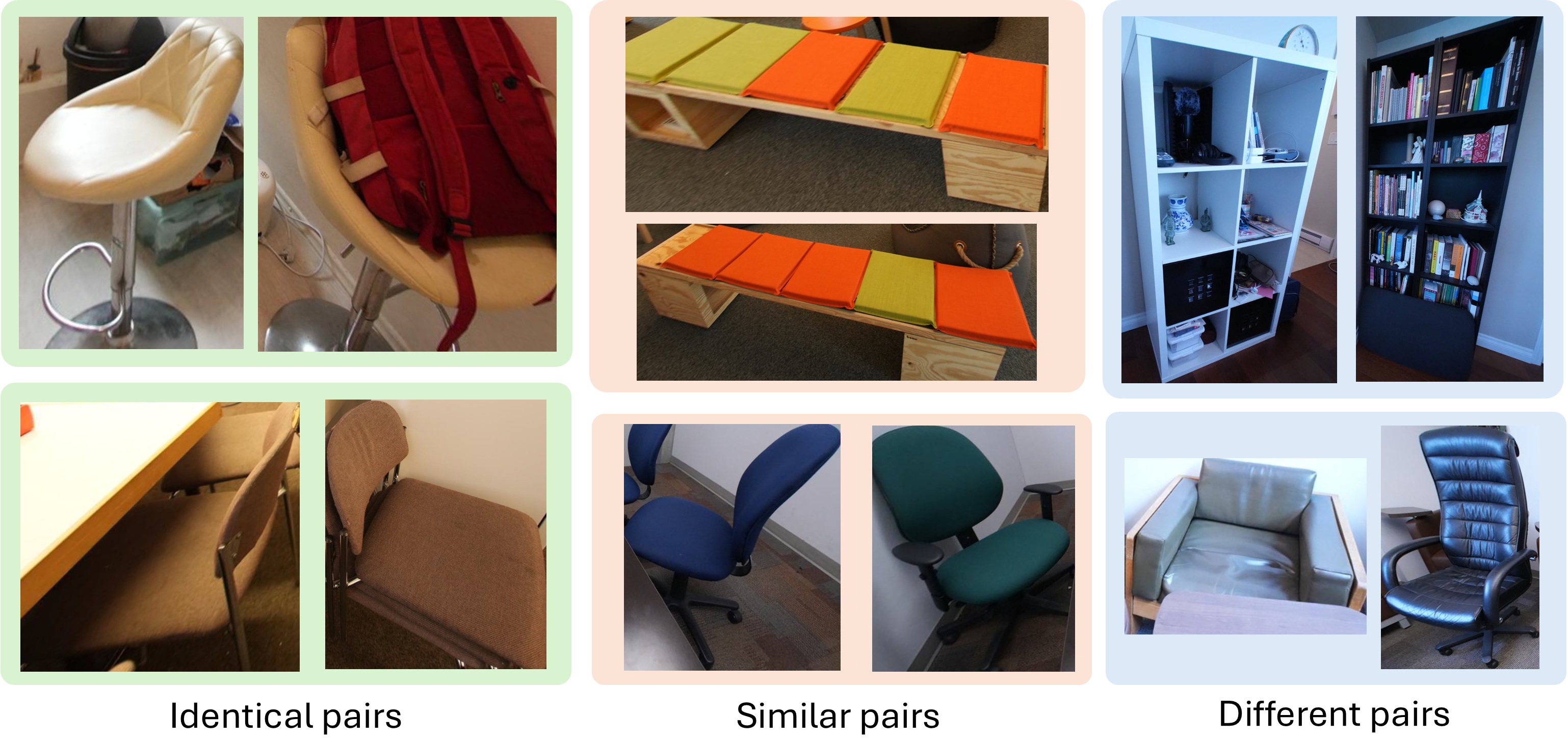}

    \caption{Examples of identical, similar and different pairs of objects from the \DatasetName{} dataset. Identical objects have the exact same shape and appearance, differing only in scene context and occlusions. Similar objects differ slightly in shape or appearance, and different objects have large differences in shape or appearance.}
    \label{fig:dataset_examples}
\end{figure*}

We introduce a dataset of identical, similar, and different object pairs based on ScanNet++ \cite{yeshwanth2023scannet++}. This dataset leverages paired 3D and RGB-D data, where dense image trajectories provide rich multiview information and 3D geometry enables consistent object association across views. During training we use only the multiview images.
We generate all intra-class object pairs, excluding structural elements (walls, floors), small items (pens, files), and objects without distinct 3D segmentation (pipes). The remaining pairs are manually annotated as identical, similar, or different, totaling \DatasetNumTotalPairs{} pairs. Identical objects must match exactly, while similar pairs differ slightly and include articulated and deformed states of objects with at least $90\%$ shape overlap. Dataset statistics and examples are provided in Tab. \ref{tab:dataset_stats} and Fig. \ref{fig:dataset_examples}; further annotation details are in the supplementary.

\section{Experiments}
\label{sec:results}

\paragraph{Evaluation Setting and Metrics}
We evaluate lookalike object detection on the \DatasetName{} validation set corresponding to the ScanNet++ validation set, as the ScanNet++ test set is hidden. We evaluate on both ground truth 3D instances and those predicted by MaskClustering \cite{yan2024maskclustering}. To address dataset imbalance, we report Intersection-over-Union (IoU) for identical, similar, and different categories, alongside the overall IoU.

\subsection{Baselines}
We compare our method against state-of-the-art 2D and 3D matching baselines, which leverage diverse indoor objects and fine-grained details. Re-identification methods are not directly comparable as they require large class-specific datasets, whereas our setting involves over 100 object classes.

2D matching methods include Telling Left from Right (TLFR) \cite{zhang2024telling} which combines Stable Diffusion \cite{rombach2022high} and DINOv2 \cite{oquab2023dinov2} features, SuperGlue \cite{sarlin2020superglue} which predicts pixel matches using a graph neural network, Mast3r \cite{leroy2024grounding} which predicts pixel features using decoders with cross attention, and Aspanformer \cite{chen2022aspanformer} which uses a combination of CNN and transformers to extract image features. 
3D matching methods include Lepard \cite{li2022lepard} and Predator \cite{huang2021predator}. \OURS{} uses semantic and structural features from a pretrained DINOv2 model, which can also be used off-the-shelf to perform image matching. Hence, we include a baseline that directly compares the raw features from the last layer of DINOv2, called DINOv2-feats. 

For each method, we empirically set two equally spaced thresholds $t_1$ and $t_2$ based on the histogram of the number of matches predicted by that method, such that atleast $95\%$ of the image pairs have a number of matches falling within this range, and further keep only matches with atleast $70\%$ confidence where applicable. The thresholds $t_1$ and $t_2$ are analogous to our similarity thresholds $t_1=0.33$ and $t_2=0.66$. The 2D matching methods are originally designed to operate on single input views. For a more fair comparison with our multiview method, we adapt these methods to the multiview setting by creating $5$ pairs of images for each object pair, and evaluate using the average number of matches across these $5$ pairs.

\subsection{Comparison to State of the Art}
Quantitative results are shown in Tab. \ref{tab:main_results}. \OURS{} outperforms all baselines in both settings: with GT instances and with predicted instances, and improves upon the best performing methods by \ImprovementGTAbs{} and \ImprovementPredAbs{} overall IoU respectively.
DINO-v2 raw features primarily capture semantics but are poorly aligned across large viewpoint changes. Similarly, 2D matching methods and Lepard rely heavily on contextual cues, leading to a bias where they struggle with identical objects despite performing well on the \textit{different} class. Predator tends to over-predict matches based on semantics alone, ignoring fine-grained shape. Furthermore, these baselines lack mechanisms to effectively leverage multi-view information. In contrast, \OURS{} integrates fine-grained shape and appearance from multiple views, achieving robust performance across all three pair categories.

\begin{figure*}[p]
    \centering
    \makebox[\textwidth][c]{
        \begin{tabular}{c c c}
            \rotatebox[origin=c]{90}{\textbf{Predator}} & 
            \includegraphics[width=0.48\linewidth, height=0.17\textheight, keepaspectratio, valign=m]{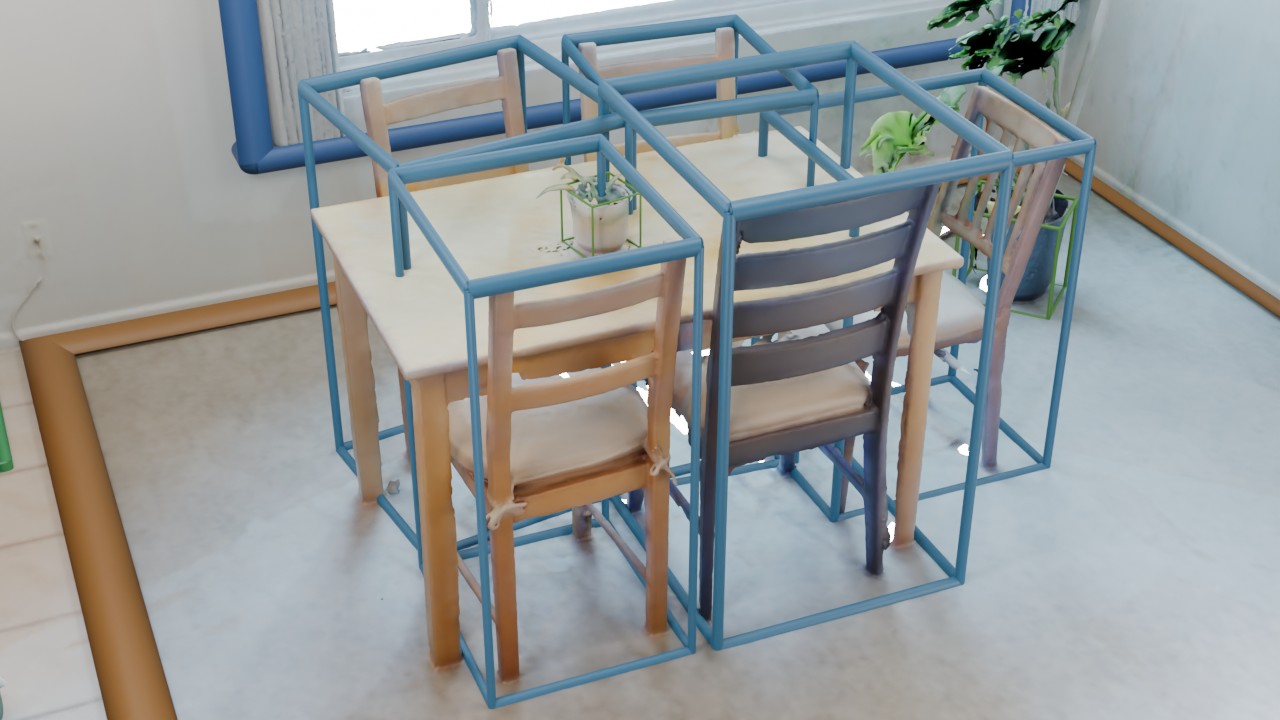} & 
            \includegraphics[width=0.48\linewidth, height=0.17\textheight, keepaspectratio, valign=m]{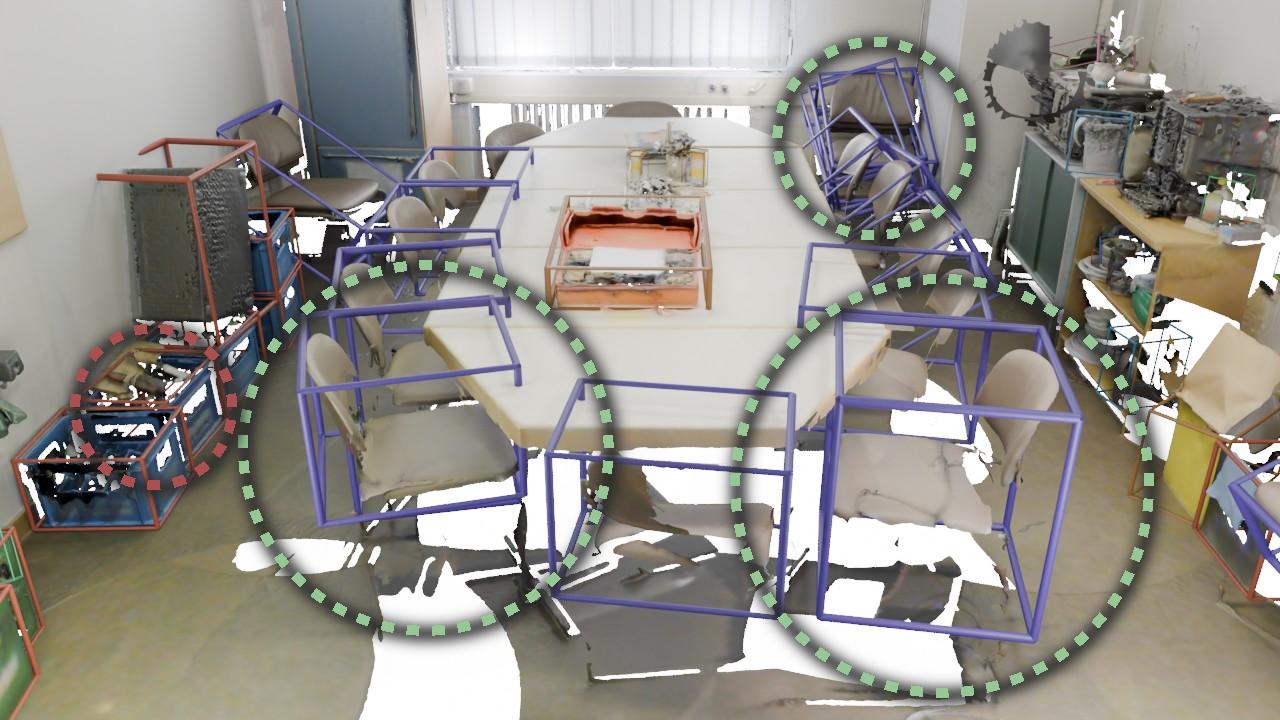} \\[10.5ex]
            
            \rotatebox[origin=c]{90}{\textbf{MASt3R}} & 
            \includegraphics[width=0.48\linewidth, height=0.17\textheight, keepaspectratio, valign=m]{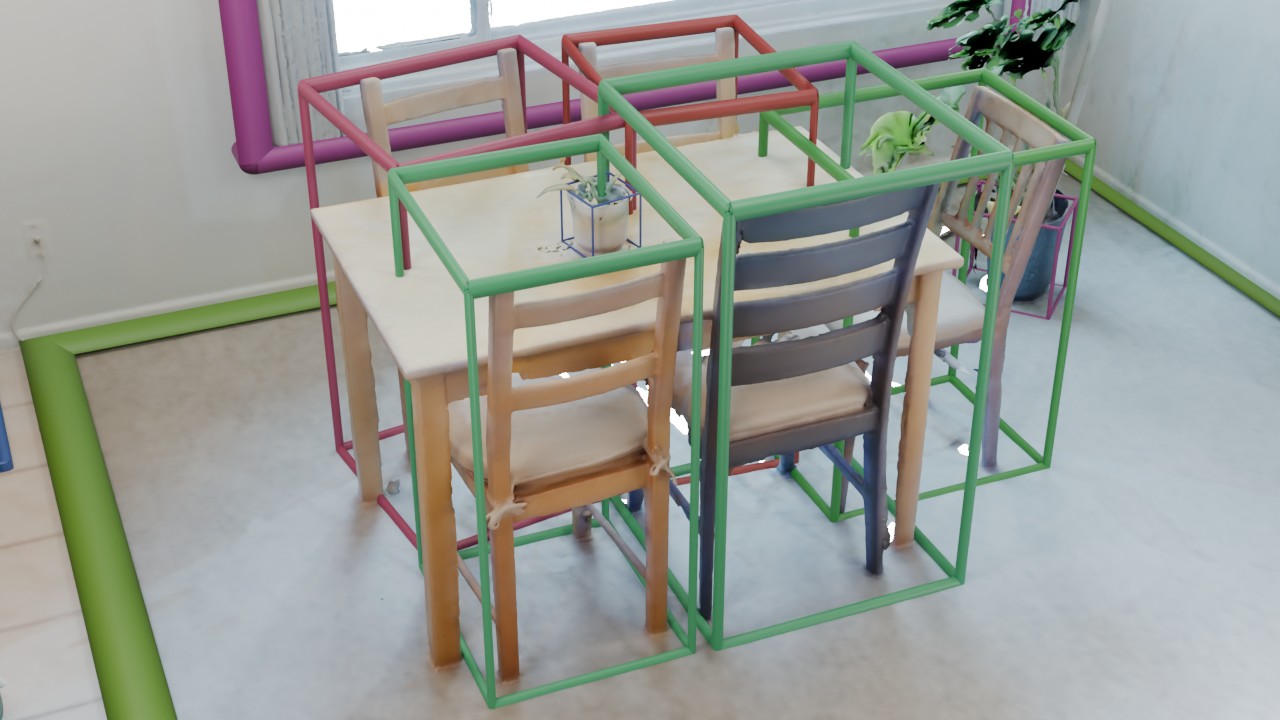} & 
            \includegraphics[width=0.48\linewidth, height=0.17\textheight, keepaspectratio, valign=m]{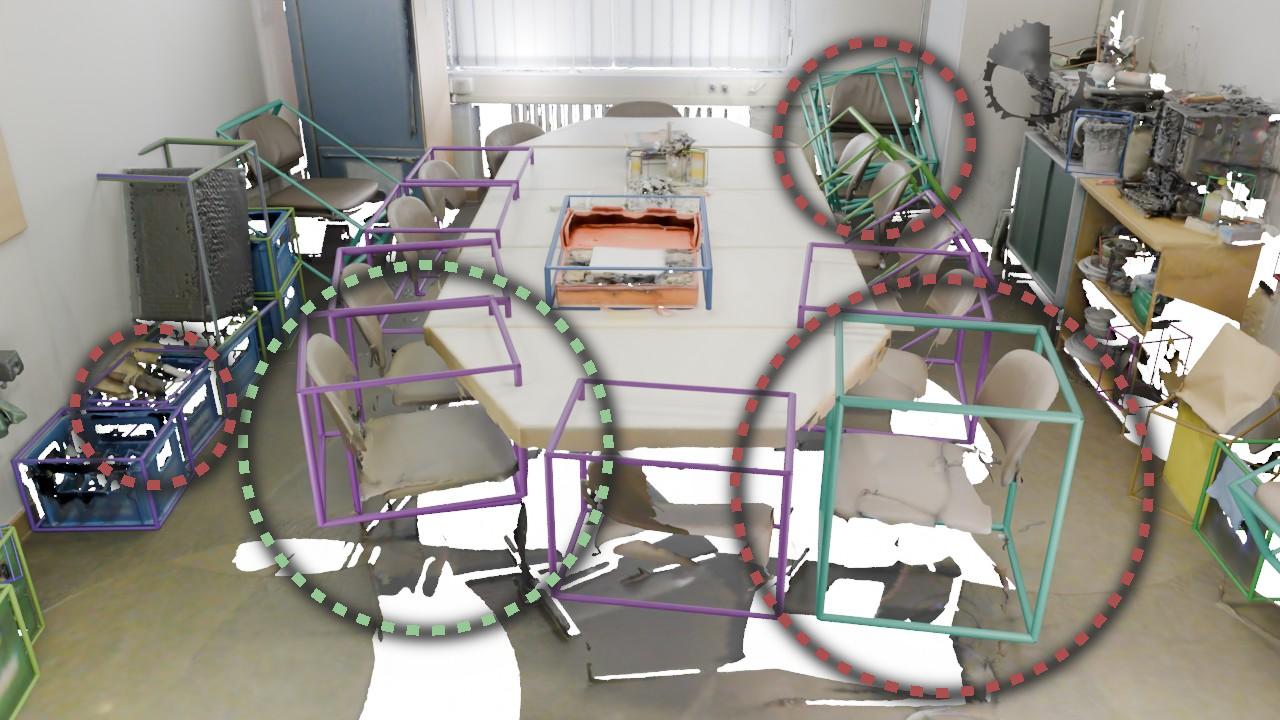} \\[10.5ex]
            
            \rotatebox[origin=c]{90}{\textbf{TLFR}} & 
            \includegraphics[width=0.48\linewidth, height=0.17\textheight, keepaspectratio, valign=m]{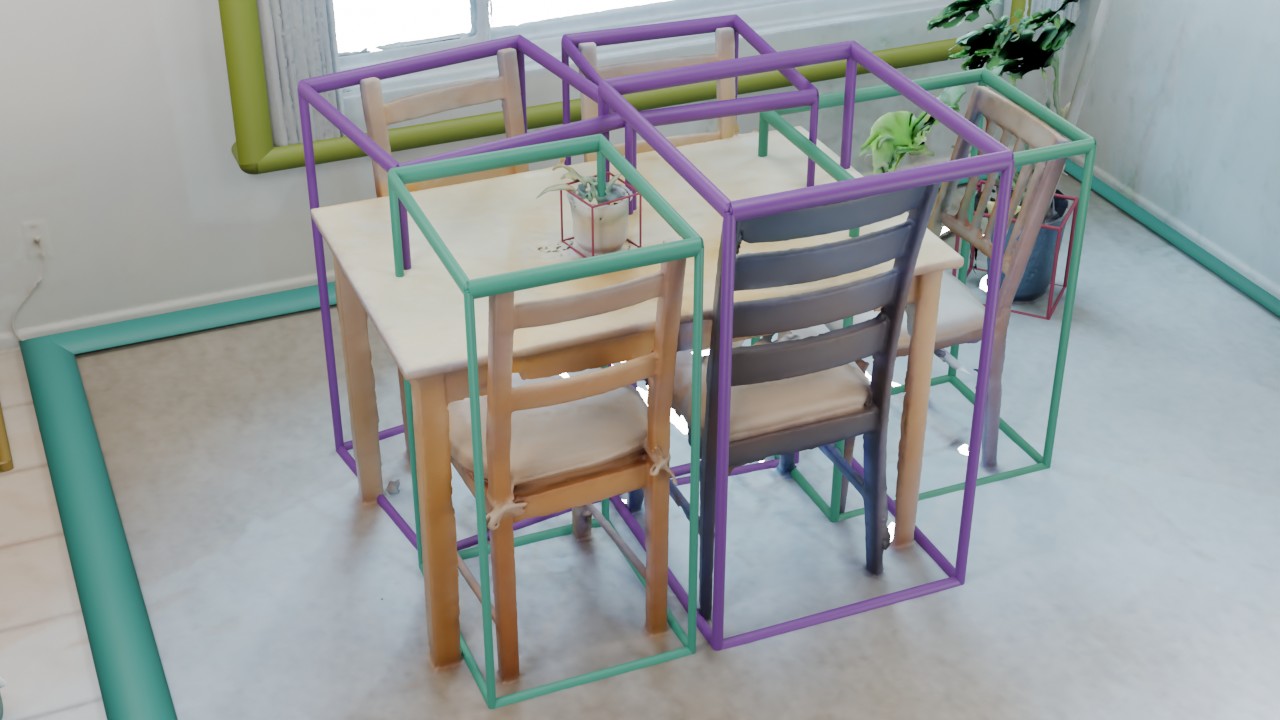} & 
            \includegraphics[width=0.48\linewidth, height=0.17\textheight, keepaspectratio, valign=m]{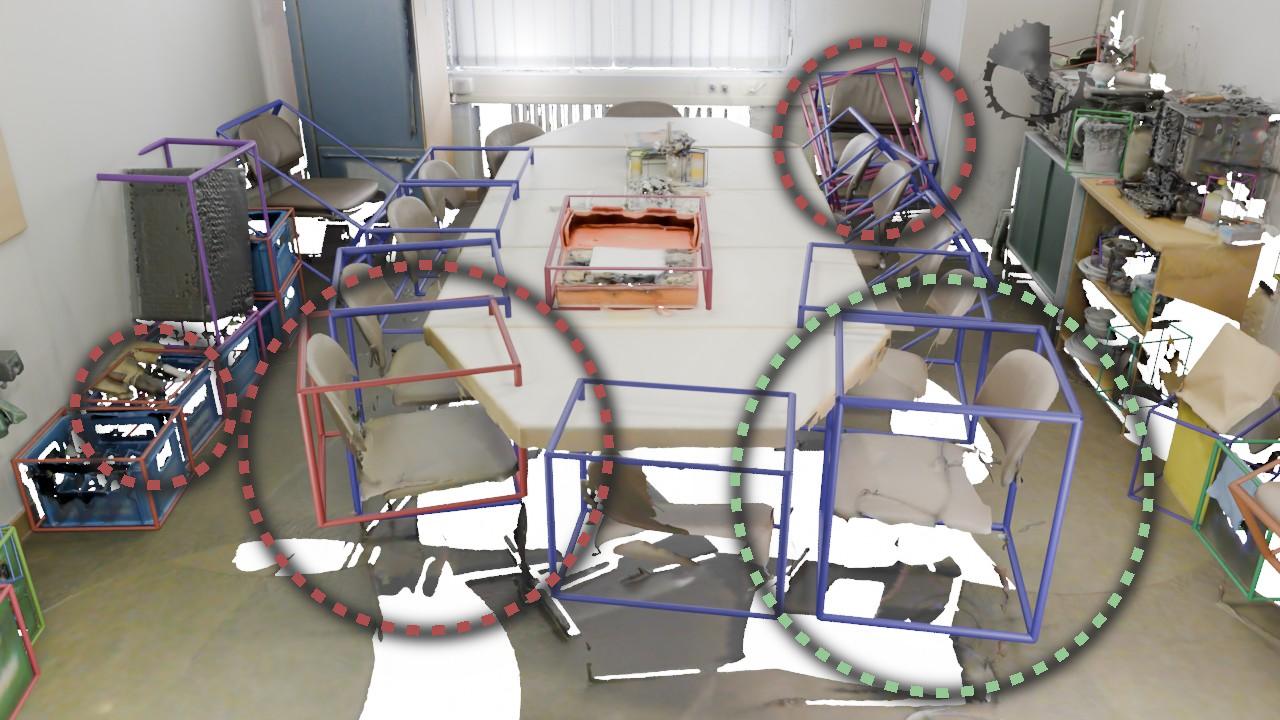} \\[10.5ex]
            
            \rotatebox[origin=c]{90}{\textbf{\OURS{} (Ours)}} & 
            \includegraphics[width=0.48\linewidth, height=0.17\textheight, keepaspectratio, valign=m]{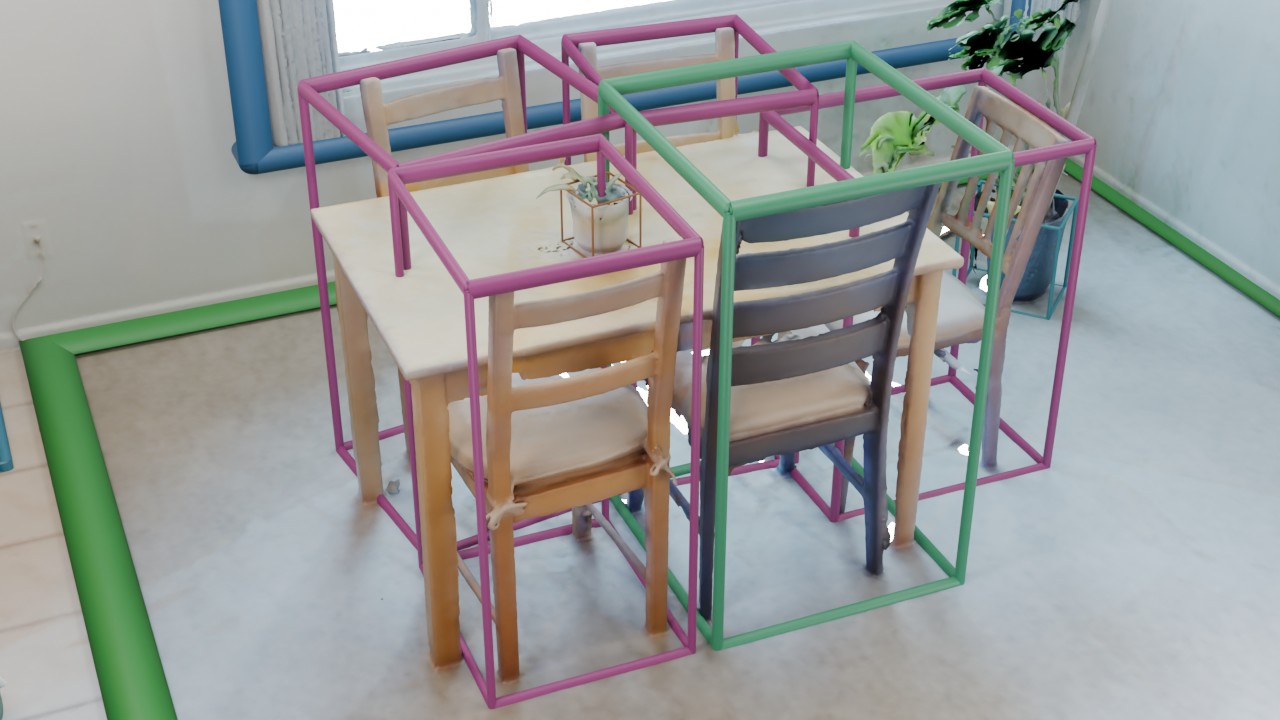} & 
            \includegraphics[width=0.48\linewidth, height=0.17\textheight, keepaspectratio, valign=m]{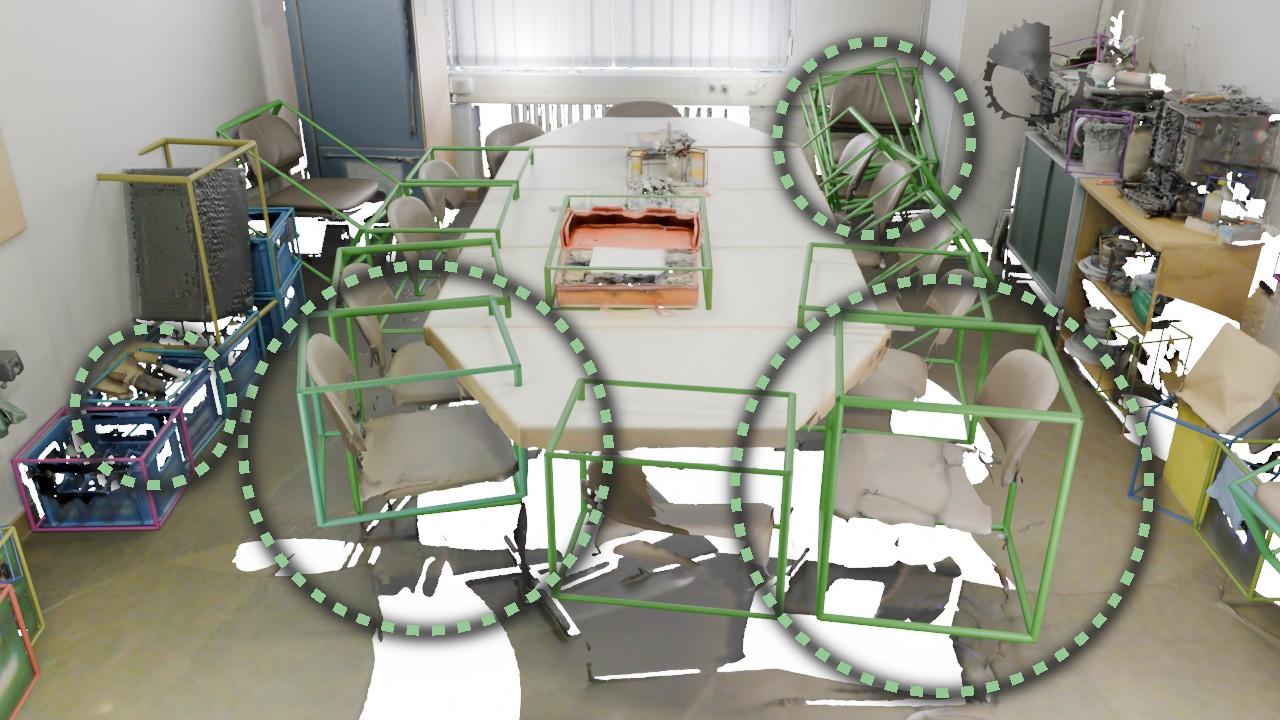} \\[10.5ex]
            
            \rotatebox[origin=c]{90}{\textbf{GT}} & 
            \includegraphics[width=0.48\linewidth, height=0.17\textheight, keepaspectratio, valign=m]{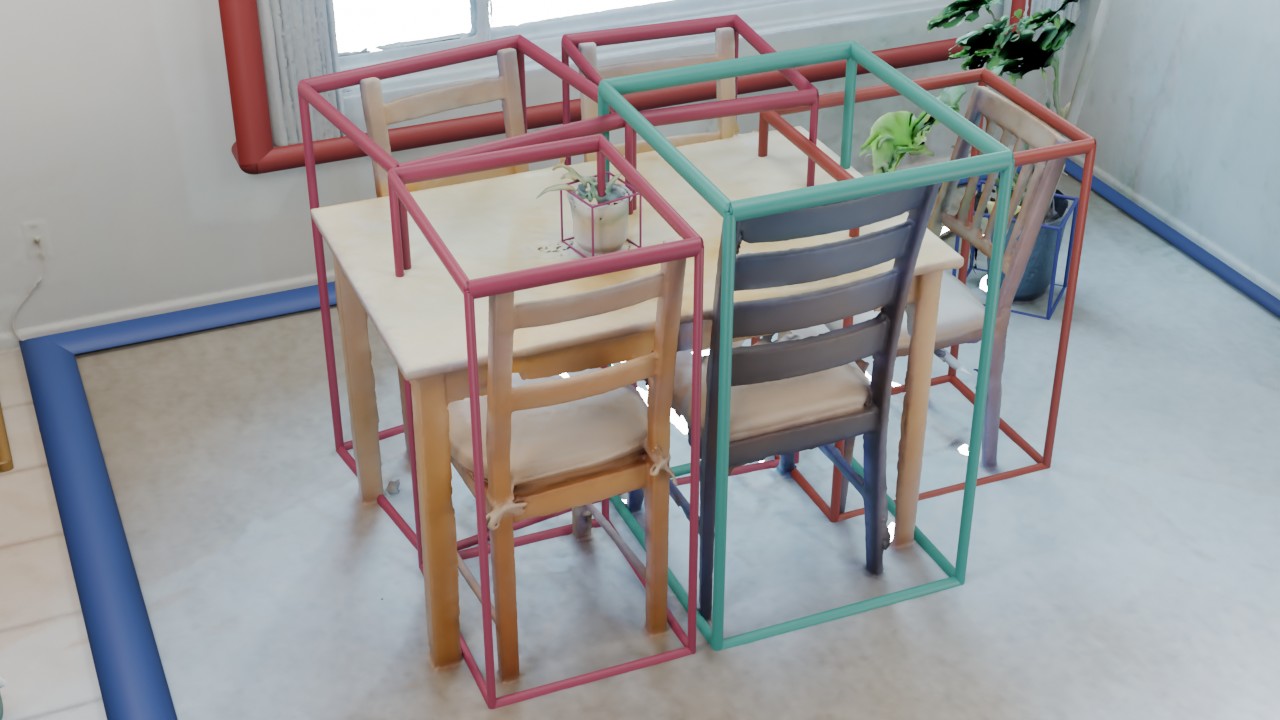} & 
            \includegraphics[width=0.48\linewidth, height=0.17\textheight, keepaspectratio, valign=m]{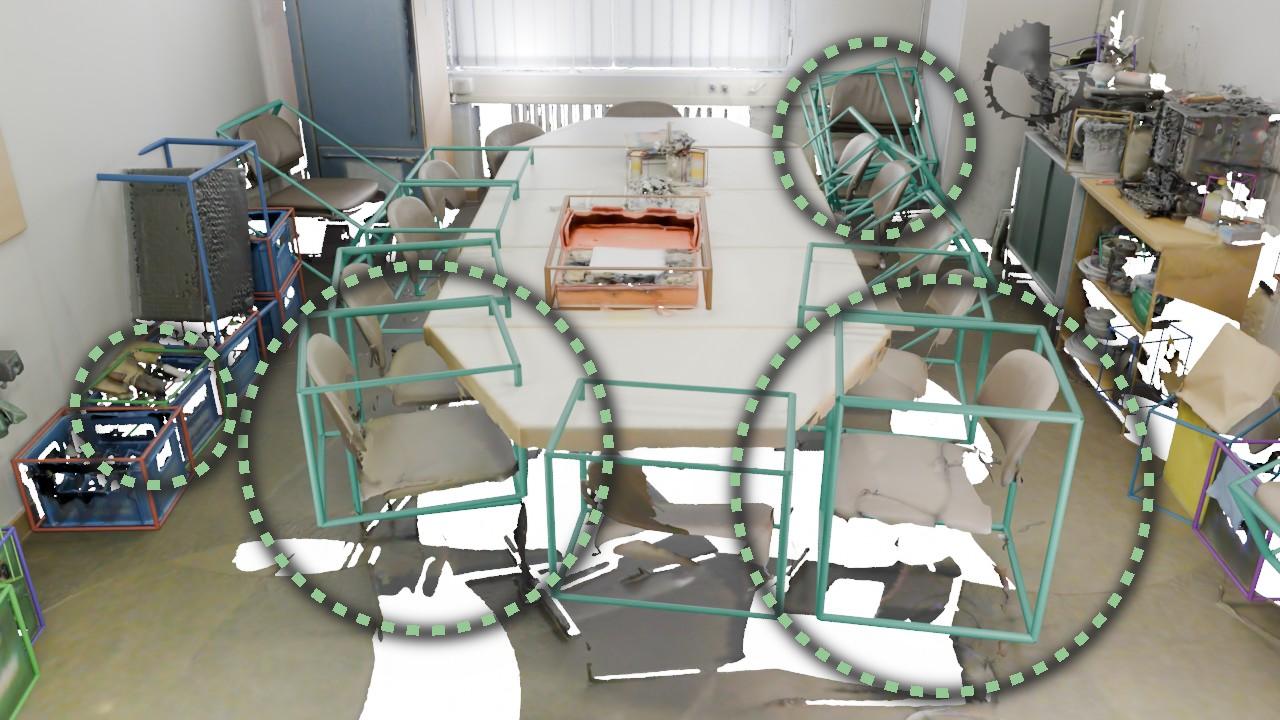} \\
        \end{tabular}
    }
    \caption{Identical object groups identified by \OURS{} and baselines. Each group is indicated by a unique color which is randomized across scenes and methods. Correct and incorrect predictions are circled in \textcolor{green!50!gray}{green} and \textcolor{red!50!gray}{red}, respectively. Unlike baselines that frequently misclassify or over-group instances, \OURS{} leverages multi-view images and fine-grained appearance for accurate object grouping.}
    \label{fig:qual_results}
\end{figure*}

Qualitative results are shown in Fig. \ref{fig:qual_results}. \OURS{} correctly recognizes identical groups of objects even in cluttered scenes with occlusion.  The baselines tend to classify these as similar or different objects, or wrongly classify objects from the same class as identical. 

\begin{table*}[t]
\centering
\label{tab:main_results_combined}
\begin{tabular}{@{}l @{\hspace{1.5em}} l cccc @{\hspace{2em}} cccc@{}}
& & \multicolumn{4}{c}{\textbf{GT Instances}} & \multicolumn{4}{c}{\textbf{Pred. Instances}} \\
\cmidrule(lr{2.0em}){3-6} \cmidrule(l){7-10}
& \textbf{Method} & Id. & Sim. & Diff. & Overall & Id. & Sim. & Diff. & Overall \\
\midrule
\multirow{2}{*}{3D Methods} & Lepard \cite{li2022lepard}               & 0.06 & 0.13 & \underline{0.47} & 0.22 & 0.05 & 0.10 & 0.24 & 0.13 \\
                    & Predator \cite{huang2021predator}        & \underline{0.52} & 0.06 & 0.08 & 0.22 & \underline{0.42} & 0.06 & 0.03 & \underline{0.17} \\
\midrule
\multirow{5}{*}{2D Methods} & DINOv2-feats \cite{oquab2023dinov2}      & 0.02 & \underline{0.14} & 0.37 & 0.18 & 0.03 & \underline{0.15} & \textbf{0.37} & 0.18 \\
& SuperGlue \cite{sarlin2020superglue}     & 0.03 & 0.07 & 0.44 & 0.18 & 0.11 & 0.02 & 0.09 & 0.07 \\
                    & ASpanformer \cite{chen2022aspanformer}   & 0.06 & 0.06 & 0.44 & 0.18 & 0.10 & 0.00 & \underline{0.10} & 0.07 \\
                    & Mast3r \cite{leroy2024grounding}         & 0.17 & 0.08 & \underline{0.47} & \underline{0.24} & \underline{0.26} & 0.02 & 0.02 & \underline{0.10} \\
                    & TLFR \cite{zhang2024telling}             & \underline{0.29} & \underline{0.12} & 0.33 & \underline{0.24} & 0.16 & \underline{0.08} & 0.05 & 0.09 \\
\midrule
\multirow{1}{*}{\textit{Ours}} & \OURS{} & \textbf{0.65} & \textbf{0.18} & \textbf{0.65} & \textbf{0.49} & \textbf{0.50} & \textbf{0.15} & \underline{0.29} & \textbf{0.31} \\
\bottomrule
\end{tabular}
\vspace{0.5em}
\caption{Quantitative comparison on the \DatasetName{} dataset. We compare \OURS{} against state-of-the-art baselines using IoU (\%) for Identical (Id.), Similar (Sim.), and Different (Diff.) pairs. We report results on both Ground Truth (GT) instances and Predicted (MaskClustering) to evaluate robustness to the input instance segmentation. The best results are \textbf{bolded} and the second best are \underline{underlined}. \OURS{} outperforms all baselines by a significant margin.}
\label{tab:main_results}
\end{table*}

\subsection{Applications}
\label{sec:applications}
\OURS{} outputs pairs of identical and similar objects in indoor scenes. We show how these identical and similar object pairs can be used in conjunction with existing 3D generative and understanding models to improve upon their performance on independent inputs.

\vspace{-0.25cm}
\subsubsection{3D Object Reconstruction}
\begin{figure*}[t]
    \includegraphics[width=\textwidth]{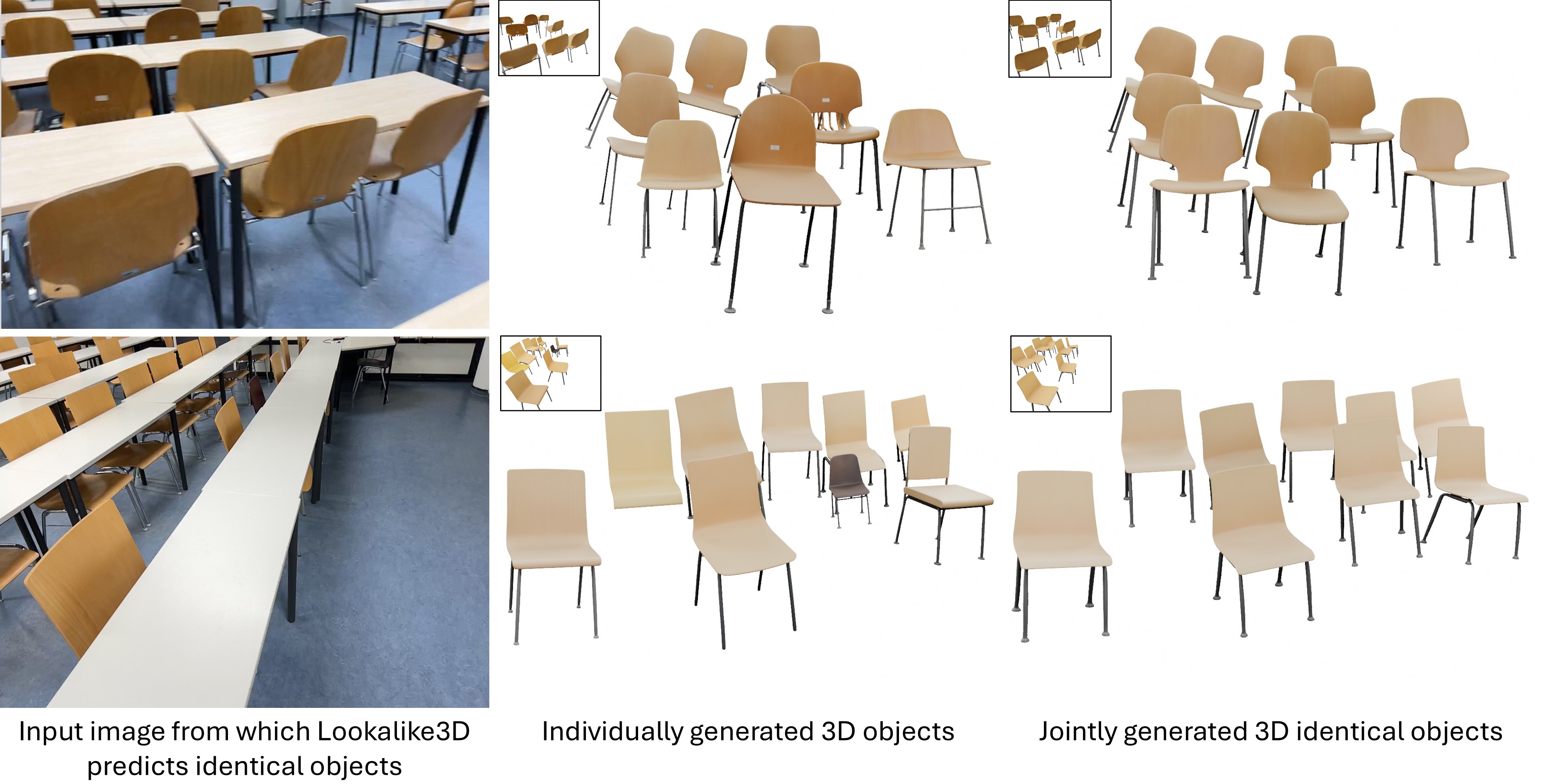}
     \caption{Joint 3D reconstruction using SAM 3D on \OURS{} outputs. While individual reconstructions are plausible, they vary significantly in shape and size, often omitting parts due to limited context. In contrast, joint prediction leverages redundant and complementary cues across instances to produce a single, consistent reconstruction. Insets show objects from their original viewpoints.}

    \label{fig:jointrecon}
\end{figure*}

We apply \OURS{} to obtain identical pairs of objects, and jointly reconstruct these pairs using SAM 3D Objects. A qualitative comparison of jointly and individually reconstructed objects is shown in Fig. \ref{fig:jointrecon}. SAM 3D objects can produce individually plausible 3D reconstructions for each input object, but these objects vary greatly in shape and scale since they have no knowledge of other identical objects in the scene. \OURS{} enables SAM 3D to create consistent 3D reconstructions of all identical objects in the scene.

\vspace{-0.25cm}

\subsubsection{Part Co-segmentation}
\begin{figure*}[t]
    \centering
    \newcommand{\subimgw}{0.31\linewidth}
    \newcommand{\columnw}{0.47\textwidth}
    
    \begin{minipage}[t]{\columnw}

        \includegraphics[width=\subimgw]{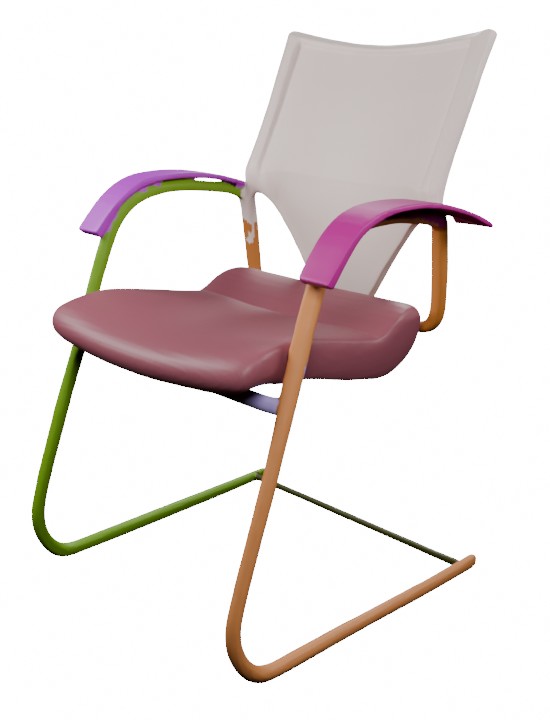} \hfill
        \includegraphics[width=\subimgw]{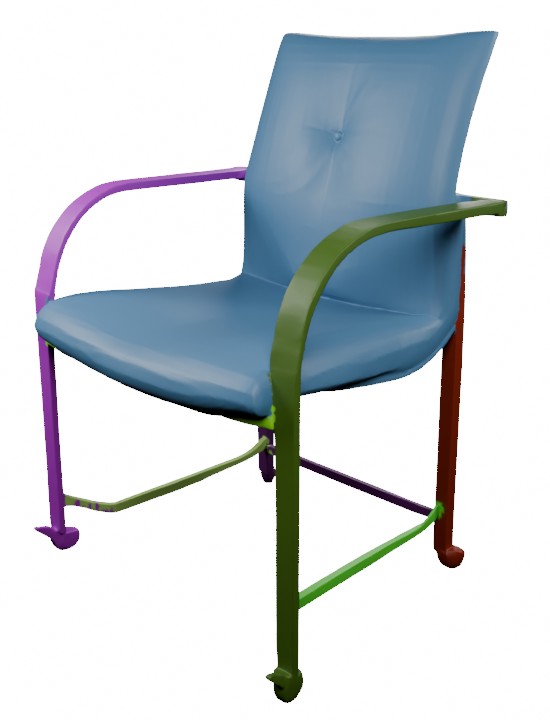} \hfill
        \includegraphics[width=\subimgw]{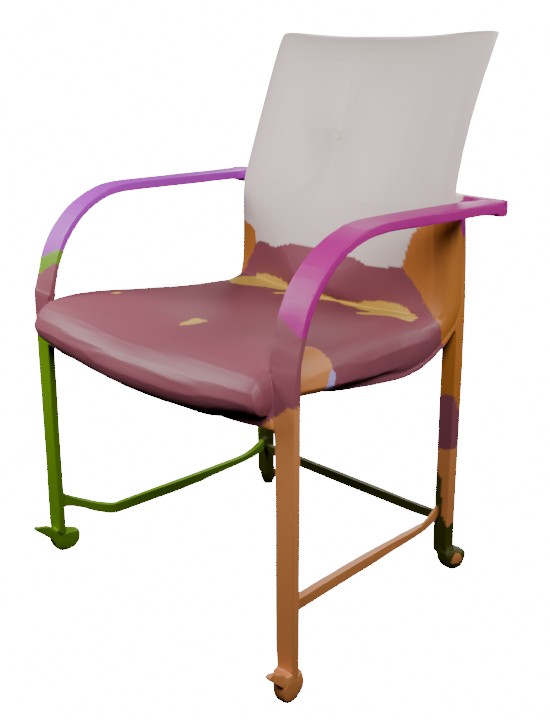}

        \vspace{1mm} %
        
        \parbox[t]{\subimgw}{\centering \scriptsize Correctly segmented} \hfill
        \parbox[t]{\subimgw}{\centering \scriptsize Undersegmented} \hfill
        \parbox[t]{\subimgw}{\centering \scriptsize Transferred segmentation}
    \end{minipage}
    \hfill 
    \hfill
    \begin{minipage}[t]{\columnw}

        \includegraphics[width=\subimgw]{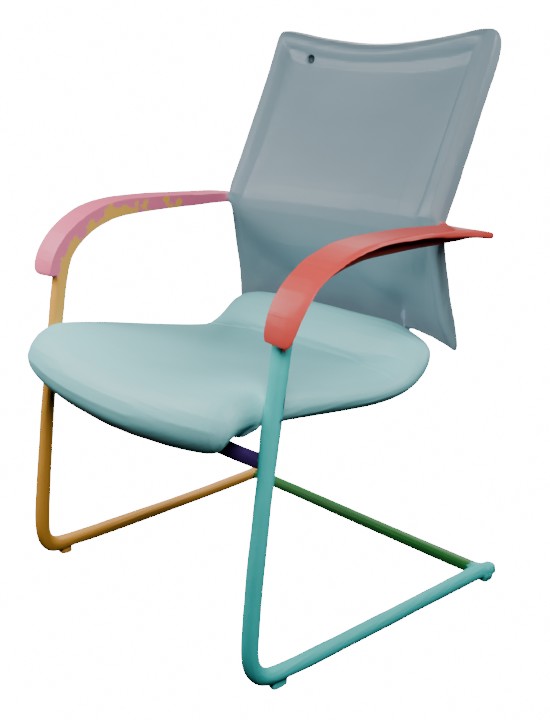} \hfill
        \includegraphics[width=\subimgw]{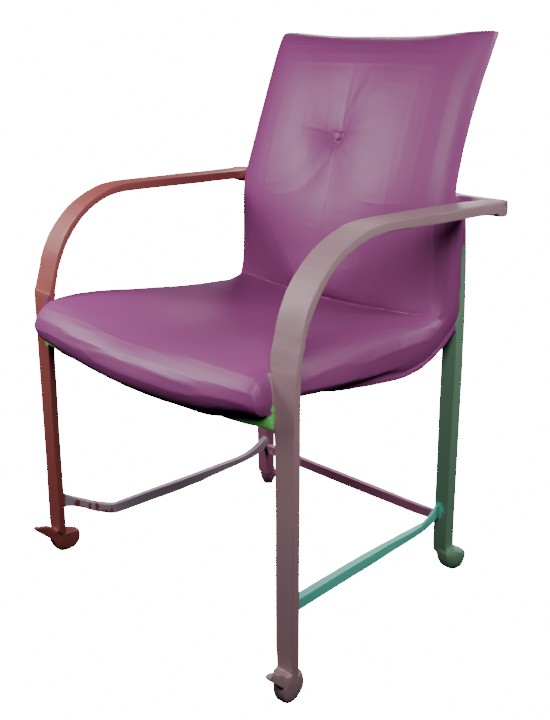} \hfill
        \includegraphics[width=\subimgw]{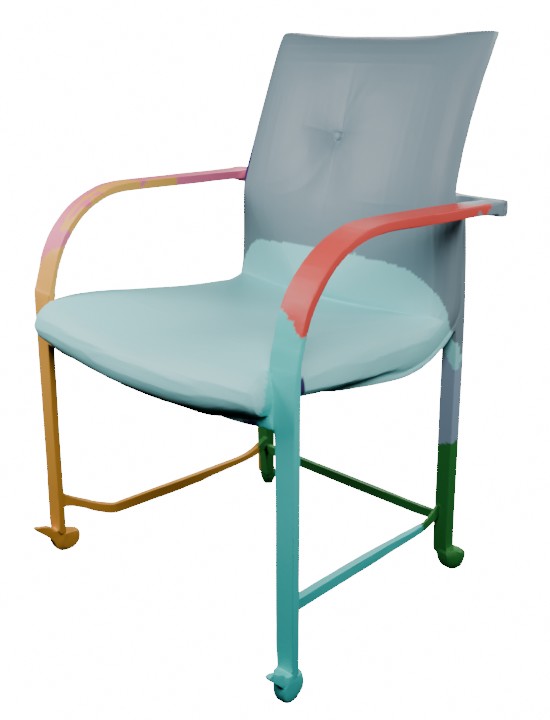}
        
        \vspace{1mm} %
        
        \parbox[t]{\subimgw}{\centering \scriptsize Correctly segmented} \hfill
        \parbox[t]{\subimgw}{\centering \scriptsize Undersegmented} \hfill
        \parbox[t]{\subimgw}{\centering \scriptsize Transferred segmentation}
    \end{minipage}

    \caption{Part segmentation transfer across similar objects. P3-SAM predicts high-quality part segmentation on objects with distinct parts, but undersegments objects with ambiguous geometry such as fused chair arms and legs. We make use of similar pairs predicted by \OURS{}, and transfer the part segmentation from the correctly segmented source object to the undersegmented target object to get more detailed and consistent segmentation.}
    \label{fig:partseg}
\end{figure*}

We apply predicted similar pairs to 3D part segmentation using P3-SAM. While P3-SAM produces high-quality parts, it often undersegments objects with ambiguous geometry, such as fused chair legs and arms. In these cases, we transfer part segmentation from a source object with clearer geometry to a similar target object. A qualitative comparison of individually and jointly obtained part segmentation is shown in Fig. \ref{fig:partseg}. By leveraging similar object pairs we can improve part segmentation on undersegmented objects.

\subsection{Ablation Studies}
We ablate the different components of our method to show
their contribution towards the full model in Tab. \ref{tab:ablation}. Each row is ablated independently from the full model by changing only one hyperparameter or model component. All ablations are run with GT instances.

\begin{table}[ht!]
    \centering
    \begin{tabular}{lcccc}
        \toprule
        Model/Data & Id. & Sim. & Diff. & Overall \\
        \midrule
        Triplet $\rightarrow $ InfoNCE loss & 0.36 & 0.14 & 0.56 & 0.35 \\
        Without alignment loss                  & 0.63 & 0.14 & 0.65 & 0.47 \\ 
        Without DINO aggr. feat.                       & 0.62 & 0.09 & 0.58 & 0.43 \\
        Only single-view attn.                               & 0.62 & 0.08 & 0.60 & 0.43 \\
        Only single-view \& object-level attn.                                & 0.62 & 0.07 & 0.63 & 0.44 \\
        Only single-view \& global attn.                                & 0.62 & 0.12 & 0.61 & 0.45 \\
        1 input view                                  & 0.60 & 0.13 & 0.55 & 0.43 \\
        3 inputs views                                 & \textbf{0.66} & 0.14 & 0.65 & 0.48 \\
        \midrule 
        Full model (\textit{Ours})                                    & 0.65 & \textbf{0.18} & \textbf{0.65} & \textbf{0.49} \\
        \bottomrule
    \end{tabular}
    \vspace{0.2cm}
    \caption{Ablation study. Performance is reported using IoU on pair type (identical, similar, different, and overall). Each ablation is compared independently to the full model, which uses triplet and alignment losses, aggregated DINO features, single-view, multiview and global attention, and 5 input views}
    \label{tab:ablation}
\end{table}

\vspace{-0.25cm}
\paragraph{Effect of alternating attention layers}
\OURS{} aggregates information across individual images, multiple views of an object, and between object pairs using an attention hierarchy (image-object-global). As shown in Tab. \ref{tab:ablation}, we ablate these levels by testing single-view only, single-view with object, and single-view with global attention. Removing any layer primarily degrades performance on the \textit{similar} class, as identifying near-identical objects requires comparing minor details across all available views of both instances.
\vspace{-0.25cm}
\paragraph{Effect of each loss function}
Our model utilizes a triplet loss to compare distances between identical, similar, and different object pairs, alongside a similarity score alignment loss to ground predictions relative to classification thresholds. As shown in Tab. \ref{tab:ablation}, we compare this to InfoNCE \cite{oord2018representation} and an ablation without alignment. Unlike InfoNCE, which lacks explicit margins and causes class overlap, the triplet loss effectively separates all three categories. The score alignment loss further improves classification by pushing $\id{}$ and $\diff{}$ scores into their respective bins, which indirectly aligns the $\simi{}$ category.
\vspace{-0.25cm}
\paragraph{Effect of increasing input views}
\OURS{} leverages multiple views of the input object pair to extract fine-grained shape and appearance information for pair classification. Increasing the number of views from 1 $\rightarrow$ 3 $\rightarrow$ 5 views increases IoU, particularly on the similar category as shown in Tab. \ref{tab:ablation}.
\vspace{-0.25cm}
\paragraph{Effectiveness of aggregated DINOv2 features}
DINOv2 provides robust fine-grained structural features from its intermediate layers \cite{amir2021deep} which are required to tell apart different objects. Removing intermediate layer features and using only the last-layer semantic features reduces performance on the similar and different classes, as shown in Tab. \ref{tab:ablation}.

\vspace{-0.25cm}
\paragraph{Limitations}
While our method can detect identical and similar objects, in long-tail settings (e.g., $>100$ object classes), its performance on small objects and objects far-off from the camera is limited by the patch resolution of the DINOv2 features. Since our method uses information from multiview images, poor camera coverage can limit performance on objects with small distinguishing details.

\section{Conclusion}
\label{sec:conclusion}
We introduced lookalike object detection in indoor scenes, the \OURS{} model for similarity learning, and the \DatasetName{} dataset. Our method leverages complementary geometry and appearance cues from repeated instances to improve joint 3D reconstruction and part co-segmentation. We hope \OURS{} inspires further research into object-level scene understanding.

\section*{Acknowledgments}
This project was supported by the ERC Starting Grant SpatialSem (101076253).

\bibliographystyle{splncs04}
\bibliography{main}

@article{steiner2026rescene4d,
  title={ReScene4D: Temporally Consistent Semantic Instance Segmentation of Evolving Indoor 3D Scenes},
  author={Steiner, Emily and Zheng, Jianhao and Howard-Jenkins, Henry and Xie, Chris and Armeni, Iro},
  journal={arXiv preprint arXiv:2601.11508},
  year={2026}
}

@inproceedings{wald2019rio,
  title={Rio: 3d object instance re-localization in changing indoor environments},
  author={Wald, Johanna and Avetisyan, Armen and Navab, Nassir and Tombari, Federico and Nie{\ss}ner, Matthias},
  booktitle={Proceedings of the IEEE/CVF International Conference on Computer Vision},
  pages={7658--7667},
  year={2019}
}

@inproceedings{therien2024object,
  title={Object re-identification from point clouds},
  author={Th{\'e}rien, Benjamin and Huang, Chengjie and Chow, Adrian and Czarnecki, Krzysztof},
  booktitle={Proceedings of the IEEE/CVF Winter Conference on Applications of Computer Vision},
  pages={8377--8388},
  year={2024}
}

@inproceedings{weinzaepfel2023croco,
  title={Croco v2: Improved cross-view completion pre-training for stereo matching and optical flow},
  author={Weinzaepfel, Philippe and Lucas, Thomas and Leroy, Vincent and Cabon, Yohann and Arora, Vaibhav and Br{\'e}gier, Romain and Csurka, Gabriela and Antsfeld, Leonid and Chidlovskii, Boris and Revaud, J{\'e}r{\^o}me},
  booktitle={Proceedings of the IEEE/CVF International Conference on Computer Vision},
  pages={17969--17980},
  year={2023}
}

@inproceedings{an2025cross,
  title={Cross-view completion models are zero-shot correspondence estimators},
  author={An, Honggyu and Kim, Jin Hyeon and Park, Seonghoon and Jung, Jaewoo and Han, Jisang and Hong, Sunghwan and Kim, Seungryong},
  booktitle={Proceedings of the Computer Vision and Pattern Recognition Conference},
  pages={1103--1115},
  year={2025}
}

@article{qi2022high,
  title={High-quality entity segmentation},
  author={Qi, Lu and Kuen, Jason and Guo, Weidong and Shen, Tiancheng and Gu, Jiuxiang and Jia, Jiaya and Lin, Zhe and Yang, Ming-Hsuan},
  journal={arXiv preprint arXiv:2211.05776},
  year={2022}
}

@article{lin2025depth,
  title={Depth anything 3: Recovering the visual space from any views},
  author={Lin, Haotong and Chen, Sili and Liew, Junhao and Chen, Donny Y and Li, Zhenyu and Shi, Guang and Feng, Jiashi and Kang, Bingyi},
  journal={arXiv preprint arXiv:2511.10647},
  year={2025}
}

@inproceedings{wang2022matchformer,
  title={Matchformer: Interleaving attention in transformers for feature matching},
  author={Wang, Qing and Zhang, Jiaming and Yang, Kailun and Peng, Kunyu and Stiefelhagen, Rainer},
  booktitle={Proceedings of the Asian conference on computer vision},
  pages={2746--2762},
  year={2022}
}

@article{ravi2024sam,
  title={Sam 2: Segment anything in images and videos},
  author={Ravi, Nikhila and Gabeur, Valentin and Hu, Yuan-Ting and Hu, Ronghang and Ryali, Chaitanya and Ma, Tengyu and Khedr, Haitham and R{\"a}dle, Roman and Rolland, Chloe and Gustafson, Laura and others},
  journal={arXiv preprint arXiv:2408.00714},
  year={2024}
}

@inproceedings{huang2025generalizable,
  title={Generalizable object re-identification via visual in-context prompting},
  author={Huang, Zhizhong and Liu, Xiaoming},
  booktitle={Proceedings of the IEEE/CVF International Conference on Computer Vision},
  pages={22539--22550},
  year={2025}
}

@inproceedings{wojke2018deep,
  title={Deep cosine metric learning for person re-identification},
  author={Wojke, Nicolai and Bewley, Alex},
  booktitle={2018 IEEE winter conference on applications of computer vision (WACV)},
  pages={748--756},
  year={2018},
  organization={IEEE}
}

@article{hermans2017defense,
  title={In defense of the triplet loss for person re-identification},
  author={Hermans, Alexander and Beyer, Lucas and Leibe, Bastian},
  journal={arXiv preprint arXiv:1703.07737},
  year={2017}
}

@inproceedings{zheng2015scalable,
  title={Scalable person re-identification: A benchmark},
  author={Zheng, Liang and Shen, Liyue and Tian, Lu and Wang, Shengjin and Wang, Jingdong and Tian, Qi},
  booktitle={Proceedings of the IEEE international conference on computer vision},
  pages={1116--1124},
  year={2015}
}

@article{lu2025clip,
  title={CLIP-SENet: CLIP-based semantic enhancement network for vehicle Re-identification},
  author={Lu, Liping and Fu, Zihao and Chu, Duanfeng and Wang, Wei and Xu, Bingrong},
  journal={IEEE Transactions on Intelligent Transportation Systems},
  volume={27},
  number={1},
  pages={1267--1278},
  year={2025},
  publisher={IEEE}
}

@inproceedings{meng2020parsing,
  title={Parsing-based view-aware embedding network for vehicle re-identification},
  author={Meng, Dechao and Li, Liang and Liu, Xuejing and Li, Yadong and Yang, Shijie and Zha, Zheng-Jun and Gao, Xingyu and Wang, Shuhui and Huang, Qingming},
  booktitle={Proceedings of the IEEE/CVF conference on computer vision and pattern recognition},
  pages={7103--7112},
  year={2020}
}

@inproceedings{li2021self,
  title={Self-supervised geometric features discovery via interpretable attention for vehicle re-identification and beyond},
  author={Li, Ming and Huang, Xinming and Zhang, Ziming},
  booktitle={Proceedings of the IEEE/CVF international conference on computer vision},
  pages={194--204},
  year={2021}
}

@inproceedings{liu2016deepvehicleid,
  title={Deep Relative Distance Learning: Tell the Difference Between Similar Vehicles},
  author={Liu, Hongye and Tian, Yonghong and Wang, Yaowei and Pang, Lu and Huang, Tiejun},
  booktitle={Proceedings of the IEEE Conference on Computer Vision and Pattern Recognition},
  pages={2167--2175},
  year={2016}
}

@inproceedings{liu2016deep,
  title={A deep learning-based approach to progressive vehicle re-identification for urban surveillance},
  author={Liu, Xinchen and Liu, Wu and Mei, Tao and Ma, Huadong},
  booktitle={European conference on computer vision},
  pages={869--884},
  year={2016},
  organization={Springer}
}

@inproceedings{schroff2015facenet,
  title={Facenet: A unified embedding for face recognition and clustering},
  author={Schroff, Florian and Kalenichenko, Dmitry and Philbin, James},
  booktitle={Proceedings of the IEEE conference on computer vision and pattern recognition},
  pages={815--823},
  year={2015}
}

@article{amir2021deep,
  title={Deep vit features as dense visual descriptors},
  author={Amir, Shir and Gandelsman, Yossi and Bagon, Shai and Dekel, Tali},
  journal={arXiv preprint arXiv:2112.05814},
  volume={2},
  number={3},
  pages={4},
  year={2021}
}

@article{oord2018representation,
  title={Representation learning with contrastive predictive coding},
  author={Oord, Aaron van den and Li, Yazhe and Vinyals, Oriol},
  journal={arXiv preprint arXiv:1807.03748},
  year={2018}
}

@inproceedings{rombach2022high,
  title={High-resolution image synthesis with latent diffusion models},
  author={Rombach, Robin and Blattmann, Andreas and Lorenz, Dominik and Esser, Patrick and Ommer, Bj{\"o}rn},
  booktitle={Proceedings of the IEEE/CVF conference on computer vision and pattern recognition},
  pages={10684--10695},
  year={2022}
}

@inproceedings{yan2024maskclustering,
  title={Maskclustering: View consensus based mask graph clustering for open-vocabulary 3d instance segmentation},
  author={Yan, Mi and Zhang, Jiazhao and Zhu, Yan and Wang, He},
  booktitle={Proceedings of the IEEE/CVF Conference on Computer Vision and Pattern Recognition},
  pages={28274--28284},
  year={2024}
}

@inproceedings{yeshwanth2023scannet++,
  title={Scannet++: A high-fidelity dataset of 3d indoor scenes},
  author={Yeshwanth, Chandan and Liu, Yueh-Cheng and Nie{\ss}ner, Matthias and Dai, Angela},
  booktitle={Proceedings of the IEEE/CVF International Conference on Computer Vision},
  pages={12--22},
  year={2023}
}

@inproceedings{yew2022regtr,
  title={Regtr: End-to-end point cloud correspondences with transformers},
  author={Yew, Zi Jian and Lee, Gim Hee},
  booktitle={Proceedings of the IEEE/CVF conference on computer vision and pattern recognition},
  pages={6677--6686},
  year={2022}
}

@inproceedings{wang2025vggt,
  title={Vggt: Visual geometry grounded transformer},
  author={Wang, Jianyuan and Chen, Minghao and Karaev, Nikita and Vedaldi, Andrea and Rupprecht, Christian and Novotny, David},
  booktitle={Proceedings of the Computer Vision and Pattern Recognition Conference},
  pages={5294--5306},
  year={2025}
}

@article{oquab2023dinov2,
  title={Dinov2: Learning robust visual features without supervision},
  author={Oquab, Maxime and Darcet, Timoth{\'e}e and Moutakanni, Th{\'e}o and Vo, Huy and Szafraniec, Marc and Khalidov, Vasil and Fernandez, Pierre and Haziza, Daniel and Massa, Francisco and El-Nouby, Alaaeldin and others},
  journal={arXiv preprint arXiv:2304.07193},
  year={2023}
}

@inproceedings{zhu2024living,
  title={Living scenes: Multi-object relocalization and reconstruction in changing 3d environments},
  author={Zhu, Liyuan and Huang, Shengyu and Schindler, Konrad and Armeni, Iro},
  booktitle={Proceedings of the IEEE/CVF Conference on Computer Vision and Pattern Recognition},
  pages={28014--28024},
  year={2024}
}

@article{cartillier20243d,
  title={3D Semantic MapNet: Building Maps for Multi-Object Re-Identification in 3D},
  author={Cartillier, Vincent and Jain, Neha and Essa, Irfan},
  journal={arXiv preprint arXiv:2403.13190},
  year={2024}
}

@inproceedings{li2022lepard,
  title={Lepard: Learning partial point cloud matching in rigid and deformable scenes},
  author={Li, Yang and Harada, Tatsuya},
  booktitle={Proceedings of the IEEE/CVF conference on computer vision and pattern recognition},
  pages={5554--5564},
  year={2022}
}

@inproceedings{huang2021predator,
  title={Predator: Registration of 3d point clouds with low overlap},
  author={Huang, Shengyu and Gojcic, Zan and Usvyatsov, Mikhail and Wieser, Andreas and Schindler, Konrad},
  booktitle={Proceedings of the IEEE/CVF Conference on computer vision and pattern recognition},
  pages={4267--4276},
  year={2021}
}

@inproceedings{zhang2024telling,
  title={Telling left from right: Identifying geometry-aware semantic correspondence},
  author={Zhang, Junyi and Herrmann, Charles and Hur, Junhwa and Chen, Eric and Jampani, Varun and Sun, Deqing and Yang, Ming-Hsuan},
  booktitle={Proceedings of the IEEE/CVF Conference on Computer Vision and Pattern Recognition},
  pages={3076--3085},
  year={2024}
}

@inproceedings{leroy2024grounding,
  title={Grounding image matching in 3d with mast3r},
  author={Leroy, Vincent and Cabon, Yohann and Revaud, J{\'e}r{\^o}me},
  booktitle={European conference on computer vision},
  pages={71--91},
  year={2024},
  organization={Springer}
}

@inproceedings{chen2022aspanformer,
  title={Aspanformer: Detector-free image matching with adaptive span transformer},
  author={Chen, Hongkai and Luo, Zixin and Zhou, Lei and Tian, Yurun and Zhen, Mingmin and Fang, Tian and Mckinnon, David and Tsin, Yanghai and Quan, Long},
  booktitle={European conference on computer vision},
  pages={20--36},
  year={2022},
  organization={Springer}
}

@inproceedings{sarlin2020superglue,
  title={Superglue: Learning feature matching with graph neural networks},
  author={Sarlin, Paul-Edouard and DeTone, Daniel and Malisiewicz, Tomasz and Rabinovich, Andrew},
  booktitle={Proceedings of the IEEE/CVF conference on computer vision and pattern recognition},
  pages={4938--4947},
  year={2020}
}

@article{li2025particulate,
  title={Particulate: Feed-Forward 3D Object Articulation},
  author={Li, Ruining and Yao, Yuxin and Zheng, Chuanxia and Rupprecht, Christian and Lasenby, Joan and Wu, Shangzhe and Vedaldi, Andrea},
  journal={arXiv preprint arXiv:2512.11798},
  year={2025}
}

@inproceedings{gao2025meshart,
	title={MeshArt: Generating Articulated Meshes with Structure-guided Transformers}, 
	author={Gao, Daoyi and and Siddiqui, Yawar and Li, Lei and Dai, Angela},
	booktitle={Proc. Computer Vision and Pattern Recognition (CVPR), IEEE},
	year={2025}
}

@inproceedings{chen2025freeart3d,
  title={Freeart3d: Training-free articulated object generation using 3d diffusion},
  author={Chen, Chuhao and Liu, Isabella and Wei, Xinyue and Su, Hao and Liu, Minghua},
  booktitle={Proceedings of the SIGGRAPH Asia 2025 Conference Papers},
  pages={1--13},
  year={2025}
}

@inproceedings{liu2025partfield,
  title={Partfield: Learning 3d feature fields for part segmentation and beyond},
  author={Liu, Minghua and Uy, Mikaela Angelina and Xiang, Donglai and Su, Hao and Fidler, Sanja and Sharp, Nicholas and Gao, Jun},
  booktitle={Proceedings of the IEEE/CVF International Conference on Computer Vision},
  pages={9704--9715},
  year={2025}
}

@article{ma2025p3,
  title={P3-sam: Native 3d part segmentation},
  author={Ma, Changfeng and Li, Yang and Yan, Xinhao and Xu, Jiachen and Yang, Yunhan and Wang, Chunshi and Zhao, Zibo and Guo, Yanwen and Chen, Zhuo and Guo, Chunchao},
  journal={arXiv preprint arXiv:2509.06784},
  year={2025}
}

@article{chen2025sam,
  title={Sam 3d: 3dfy anything in images},
  author={Chen, Xingyu and Chu, Fu-Jen and Gleize, Pierre and Liang, Kevin J and Sax, Alexander and Tang, Hao and Wang, Weiyao and Guo, Michelle and Hardin, Thibaut and Li, Xiang and others},
  journal={arXiv preprint arXiv:2511.16624},
  year={2025}
}

@inproceedings{xiang2025structured,
  title={Structured 3d latents for scalable and versatile 3d generation},
  author={Xiang, Jianfeng and Lv, Zelong and Xu, Sicheng and Deng, Yu and Wang, Ruicheng and Zhang, Bowen and Chen, Dong and Tong, Xin and Yang, Jiaolong},
  booktitle={Proceedings of the IEEE/CVF conference on computer vision and pattern recognition},
  pages={21469--21480},
  year={2025}
}

@article{
    xiang2025trellis2,
    title={Native and Compact Structured Latents for 3D Generation},
    author={Xiang, Jianfeng and Chen, Xiaoxue and Xu, Sicheng and Wang, Ruicheng and Lv, Zelong and Deng, Yu and Zhu, Hongyuan and Dong, Yue and Zhao, Hao and Yuan, Nicholas Jing and Yang, Jiaolong},
    journal={Tech report},
    year={2025}
}
\clearpage
\title{Supplementary Material for \OURS{}: Seeing Double in 3D}

\author{} %
\institute{}
\makeatletter
\let\@maketitle\@oldmaketitle
\makeatother
\maketitle
\section{Comparison with Single-view Baselines}
The 2D matching methods that we compare against are designed to operate in a single view setting, hence we report their single view results on the image with the most coverage, compared with our full method in Tab. \ref{tab:mv_baselines}. Single-view matching methods capture limited object information, while \OURS{} effectively combines multiview information and outperforms all baselines.

\begin{table*}[ht]
\centering
\label{tab:main_results_combined}
\begin{tabular}{@{}l @{\hspace{1.5em}} l cccc @{\hspace{2em}} cccc@{}}
\toprule
& & \multicolumn{4}{c}{\textbf{GT Instances}} & \multicolumn{4}{c}{\textbf{Pred. Instances}} \\
\cmidrule(lr{2.0em}){3-6} \cmidrule(l){7-10}
& \textbf{Method} & Id. & Sim. & Diff. & Overall & Id. & Sim. & Diff. & Overall \\
\midrule
\multirow{4}{*}{2D Methods} & SuperGlue \cite{sarlin2020superglue}     & 0.05 & 0.07 & 0.44 & 0.19 & 0.02 & 0.05 & 0.24 & 0.10 \\
                    & ASpanformer \cite{chen2022aspanformer}   & 0.07 & 0.06 & 0.43 & 0.19 & 0.04 & 0.03 & 0.24 & 0.10 \\
                    & Mast3r \cite{leroy2024grounding}         & 0.16 & 0.07 & 0.45 & 0.23 & 0.10 & 0.06 & 0.24 & 0.13 \\
                    & TLFR \cite{zhang2024telling}             & 0.25 & 0.11 & 0.33 & \underline{0.23} & 0.18 & 0.11 & 0.17 & 0.15 \\
\midrule
\textit{Ours} & \OURS{} & \textbf{0.65} & \textbf{0.18} & \textbf{0.65} & \textbf{0.49} & \textbf{0.50} & \textbf{0.15} & \textbf{0.29} & \textbf{0.31} \\
\bottomrule
\end{tabular}
\vspace{0.5em}
\caption{\OURS{} compared with 2D matching baselines evaluated on multiview images using IoU (\%) for Identical (Id.), Similar (Sim.), and Different (Diff.) pairs. We report results on both Ground Truth (GT) instances and Predicted instances to evaluate robustness. The best results are \textbf{bolded} and the second best are \underline{underlined}. The baselines improve upon their single view versions in terms of identical IoU, but degrade in overall IoU as matches are inconsistent across views.}
\label{tab:mv_baselines}
\end{table*}

\section{Dataset Details}
\subsection{Data Preparation}
We rasterize the ScanNet++ \cite{yeshwanth2023scannet++} 3D instance annotations onto the undistorted DSLR images, to obtain high quality object crops. We calculate the visibility of an object $o$ in image $I$ as the fraction of the total vertices on $o$ that are visible in $I$, and use the top 10 crops for each object duration annotation. Object pairs are created within semantic classes. Pairs are created using only the objects in the same scene.

\subsection{Annotation Interface}
\begin{figure*}[ht]
\centering
\includegraphics[width=\textwidth]{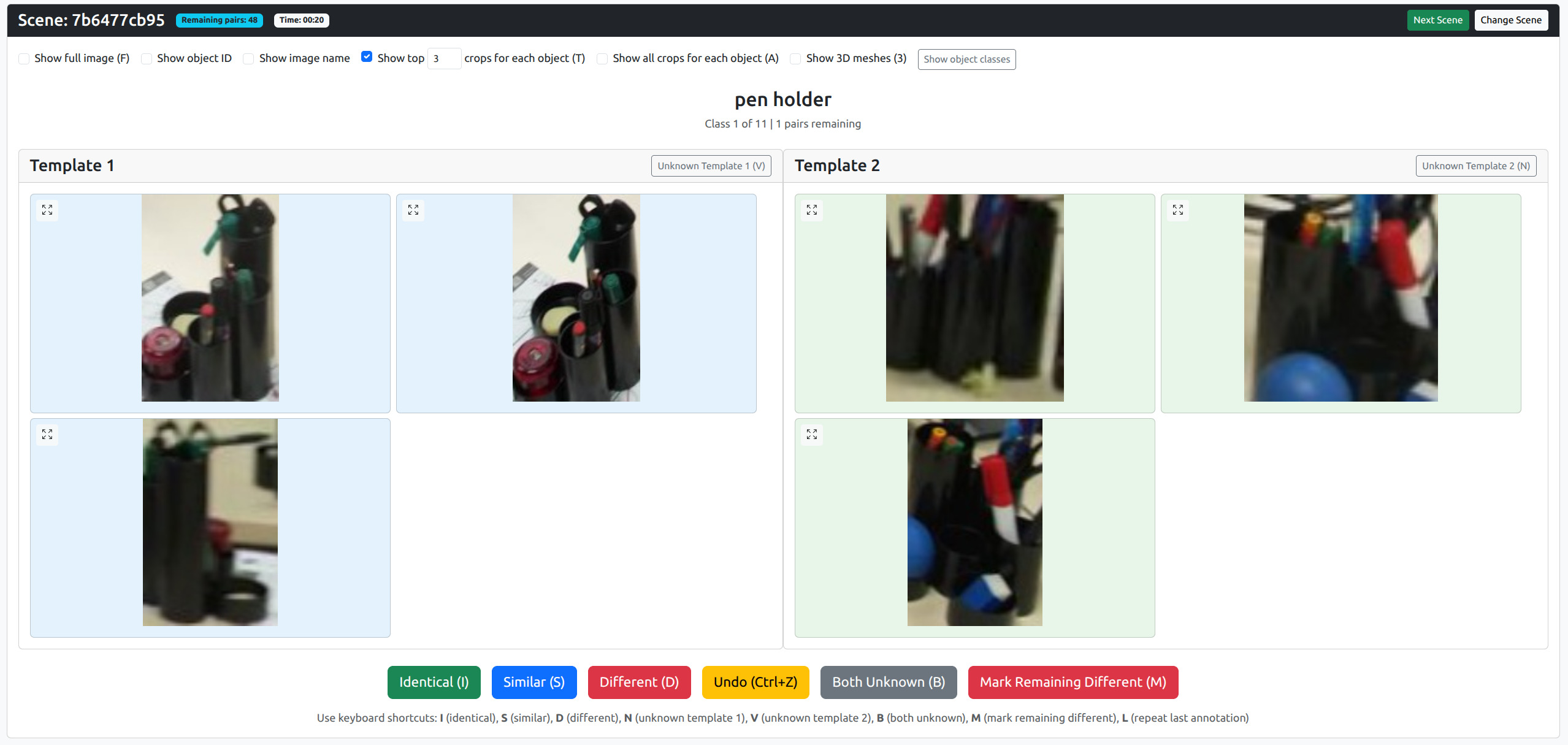}
\caption{Web interface of our annotation tool. The annotator has access to multiple views of each object as well as their full views for scene context. Here we show the top 3 views of each object by coverage.}
\label{fig:annotation_tool}
\end{figure*}

We built a web annotation interface where an annotator is shown pairs of objects and can classify each pair as identical, similar or different, and additionally select similarity types for similar pairs (described below). An example pair is shown in Fig. \ref{fig:annotation_tool}. Identical objects are iteratively grouped together; initially each object is in its own group and identical pairs are accumulated together by merging groups, which greatly reduces the time required to annotate them. Blurred or unclear pairs can be marked as unknown. The annotator can also view the full scene context of the object, which is often helpful to view repeated objects that are close together in the scene.

\begin{figure*}[ht]
\centering
\includegraphics[width=\textwidth]{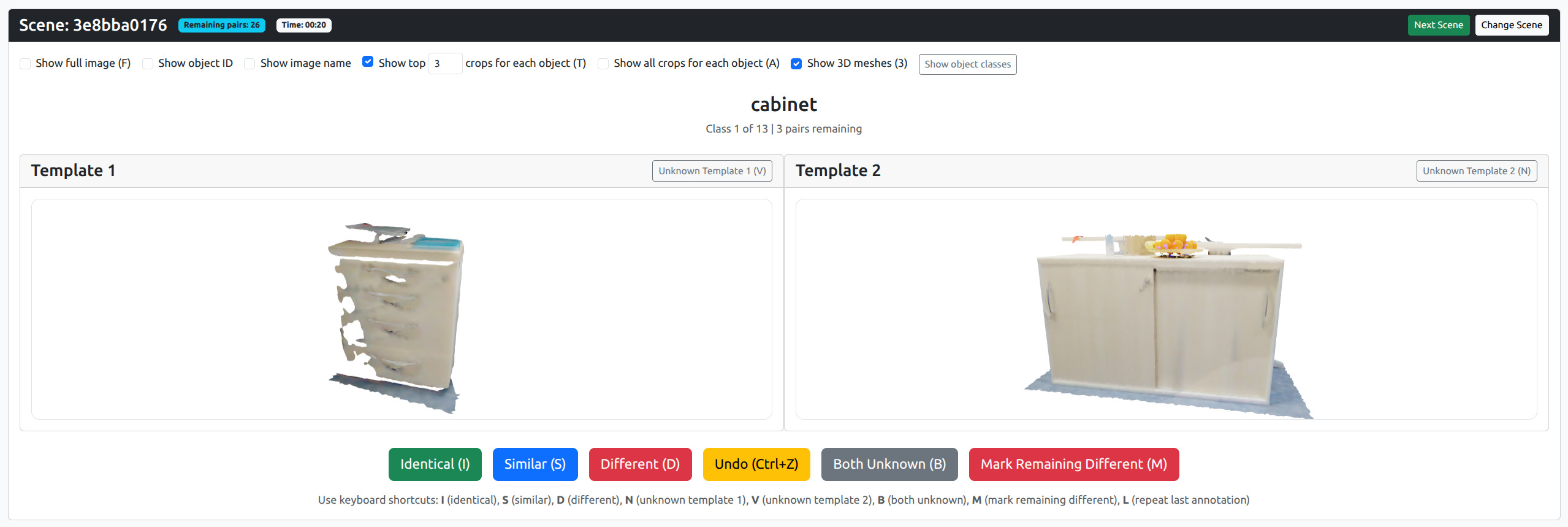}
\caption{3D view of object meshes during annotation. Note that 3D information is only available during annotation and not during model training.}
\label{fig:anno3d}
\end{figure*}

The annotator can also view the 3D mesh of the object as shown in Fig. \ref{fig:anno3d}, which is useful when the shape or dimensions of the objects are not clear from the images.

\subsection{Similarity Types}
We annotated each similar object pair with one or more \textit{similarity types}, indicating in what aspects the objects differ. Possible values for the similarity type are: 
\begin{enumerate} %
\item \textit{shape}: objects with similar shapes, and possibly different textures
\item \textit{articulated}: identical objects in different states of articulation, such as rotated office chairs and monitors adjusted to different heights
\item \textit{mirrored}: identical objects that are left-right mirrored and hence cannot be rigidly transformed to superimpose one on the other
\item \textit{deformable}: identical objects that are deformed into different 3D shapes, such as pillows and blankets
\item \textit{texture/color}: objects with identical shapes and different textures or colors.
\end{enumerate}

\subsection{Annotation Guidelines}
We ensure the quality of our annotations through the following annotation guidelines.
\textit{Identical pairs} must be unambiguously identified through the 2D images, or verified using the 3D mesh when required.
\textit{Similar pairs} must share the semantic structure, such as a pair of chairs which both have rotating wheels, or a pair of height adjustable tables at different heights, articulated table lamps, trash cans of different colors, pillows with different deformations.
\textit{Different pairs} do not have shared semantic structure and differ significantly, such as a wooden dining chair without arms, and a leather office chair with wheels and arms.
We first annotated 50 randomly selected scenes to build the annotation guidelines, and then annotated from scratch according to the guidelines to create the dataset of $856$ train and $50$ validation scenes based on ScanNet++.

\section{Implementation Details}
\subsection{Model Details}
Intermediate DINO features from layers $[1,3,5,8]$ and the last layer features are concatenated together along the feature dimension and then projected to the original feature dimension of $384$ before being passed to the alternating attention layers.

The attention layers use an internal embedding dimension of $256$, with $8$ attention heads. We add cosine positional embeddings to the patch features at each layer in our model.

\subsection{\OURS{} Training}
We apply random rotation, horizontal flip, color jitter, and crop augmentations during training. Images are padded to a square shape and then resized to the $224\times224$ input dimensions of DINO. We sample batches to maximize the number of similar and different pairs for a given anchor identical pair, and train with a batch size of at most 128 pairs.

The triplet loss requires each identical anchor pair to have a corresponding negative (similar or different) pair such that their similarities can be optimized jointly. Since each annotated identical pair does not necessarily have a similar or different pair with a common anchor object, we randomly sample \textit{different} pairs to create a more complete dataset. For an annotated identical pair of objects $(a,b)$ from semantic class $S$, we sample additional pairs $(a, c)$ and $(b, d)$ where $c,d\notin S$.

\subsection{MaskClustering Instance Segmentation}
We obtain class-agnostic instance predictions from MaskClustering, by combing Cropformer \cite{qi2022high} masks from multiview DSLR images subsampled by a factor of 5. Predicted instances are associated with GT instances based on a minimum IoU threshold on $0.5$.

\subsection{Evaluation}
We evaluate intersection-over-union (IoU) separately for each of three pair types, and over all pair types. Evaluation on GT instances allows us to separate out the accuracy of instance prediction from the accuracy of \OURS{}. 

When evaluating on predicted instances, we rasterize the predicted instance segmentation onto DSLR images and create crops as in the GT setting. GT pairs that are missing in the predictions are classified into an \textit{unknown} class, hence including the effect of instance segmentation on lookalike object detection.

\section{SAM 3D Joint Reconstruction Details}
Given identical 2D input objects, SAM 3D Objects produces 3D reconstructions that vary in shape and scale, making them inconsistent. We make use of predicted identical object pairs to jointly predict a single clean reconstruction that combines aspects of each individual object, by conditioning SAM 3D on a single image input and multiple object masks.

Specifically, we use a single RGB image $I$ containing all of the identical objects as input to SAM 3D, and the masks $M_i$ of each of these objects in $I$. Each object may have a different degree of visibility; our joint reconstruction method combines these together into a single coherent 3D object.

SAM 3D consists of a geometry model that predicts the initial dense object shape, and a texture and refinement model that refines the dense shape into a set of sparse voxels and predicts object texture.

\paragraph{Geometry Model}
The geometry model denoises a set of layout tokens into the object rotation, translation and scale, and a $16^3$ set of layout tokens densely representing the low resolution shape of the object. It consists of a sparse structure generator and a sparse structure decoder. The sparse structure generator is conditioned on $I$ and $M_i$ through cross-attention. Since the model originally conditions on a single object mask, we inject multiple masks $M_i$ of objects $o_i$ into each flow transformer layer by conditioning on each $M_i$ separately. This gives intermediate features $\{f_{l,i}\}$ corresponding to each mask in each layer $l$. We average over these features to obtain a single intermediate representation $f_{l}$ that represents all the objects $o_i$ in the canonical space, and use this as input to the $(l+1)^{th}$ layer. The final output of the sparse structure generator is a $16^3 \times 8$ feature vector representing the joint low-resolution shape of the object. This is upsampled to $64^3$ using the existing sparse structure decoder. The layout tokens are similarly decoded through the layout decoder, and the final outputs of the geometry model are a single rotation, scale and translation, and the single $64^3$ high-resolution geometry of the jointly reconstructed object.

\paragraph{Texture and Refinement Model}
The multi-mask conditioning in the texture and refinement model is carried out in a similar manner. The high-resolution $64^3$ shape is converted to sparse coordinates which are denoised to SLAT (structured latent) features. The input to the SLAT sampler is a set of noise tokens at these sparse coordinates. The SLAT sampler is then conditioned on $M_i$ and $I$, by injecting these as conditions into every layer of the sparse flow transformer and averaging the intermediate outputs. The output is a single set of SLAT features. These SLAT features are decoded with the SLAT decoder to obtain the mesh of the jointly reconstructed object.

In this way, we adapt SAM 3D to generate a single 3D object representing a combination of the objects $o_i$, effectively combining appearance and shape cues from identical objects with varying degrees of occlusion.

We use the median scale of the individually predicted objects as the scale of the jointly reconstructed object, and apply the rotation and translation of the individual objects to the joint object to transform them in the scene. Hence, the outputs from our joint reconstruction do not vary either in shape or scale, while improving upon the 3D geometry of the individual reconstructions.

\end{document}